\documentclass[final,1p,times]{elsarticle}

\journal{Journal Name}

\usepackage{amsmath,amsfonts,bm}

\def\eqref#1{equation~\ref{#1}}

\def\1{\bm{1}}

\DeclareMathAlphabet{\mathsfit}{\encodingdefault}{\sfdefault}{m}{sl}
\SetMathAlphabet{\mathsfit}{bold}{\encodingdefault}{\sfdefault}{bx}{n}

\usepackage{graphicx}
\usepackage{booktabs}
\usepackage{multirow}
\usepackage{subcaption}
\usepackage{url}
\usepackage[hidelinks]{hyperref}
\usepackage{subcaption}
\begin{document}

\begin{frontmatter}

\title{Inducing Permutation Invariant Priors in Bayesian Optimization for Carbon Capture and Storage Applications}

\author[ntua]{Sofianos Panagiotis Fotias\corref{cor1}}
\ead{sfotias@metal.ntua.gr}
\author[ntua]{Vassilis Gaganis}
\ead{vgaganis@metal.ntua.gr}

\cortext[cor1]{Corresponding author}

\affiliation[ntua]{
  organization={School of Mining and Metallurgical Engineering, National Technical University of Athens},
  city={Athens},
  postcode={15773},
  country={Greece}
}

\begin{abstract}
Bayesian Optimization is an iterative method, tailored to optimizing expensive black box objective functions. Surrogate models like Gaussian Processes, which are the gold standard in Bayesian Optimization, can be inefficient for inputs with permutation symmetries, as the most common kernels employed are better suited for vector inputs rather than unordered sets of items. Motivated by this issue, we turn to permutation invariant Bayesian Optimization for well placement in Carbon Capture and Storage projects. The high fidelity black box simulator is instructed to operate wells under group control, giving rise to permutation symmetries within injector and producer groups that cannot be exploited with standard GP kernels. In this work, our main contribution is a novel Gaussian Process kernel (GP-Perm) that encodes permutation invariance by comparing sets through a stable divergence between their induced empirical representations, and can be combined with standard kernels for additional vector-valued inputs. As a learned invariant baseline, we also consider a Deep Kernel Learning model (DKL-DS) using the Deep Sets architecture to learn a permutation-invariant embedding. We evaluate the proposed methodology across 8 use cases, comprising seven synthetic benchmarks and one realistic CCS case study (Johansen formation).
\end{abstract}


\end{frontmatter}

\section{Introduction}
\label{introduction}
Optimizing the design and operation of complex physical systems is a critical challenge in science and engineering, from drug discovery to aerospace design. A salient example is the strategic deployment of Carbon Capture and Storage (CCS), a key technology for mitigating climate change \citep{bui2021role}. Optimizing a CCS project, for instance, involves determining the locations of injection and production wells and their operational settings to maximize the total volume of securely stored CO$_2$, while simultaneously minimizing operational costs and long-term leakage risks. Finding such an optimum requires navigating a challenging design space where each evaluation entails a computationally expensive, high-fidelity numerical simulation \citep{ismail2023carbon}. Such problems are prime candidates for Bayesian Optimization (BO), a sample-efficient methodology for global optimization of black-box functions that has proven effective in a wide range of scientific applications \citep{frazier2018tutorial}.

\begin{figure}[t!]
\begin{center}
\includegraphics[width=0.75\linewidth,keepaspectratio]{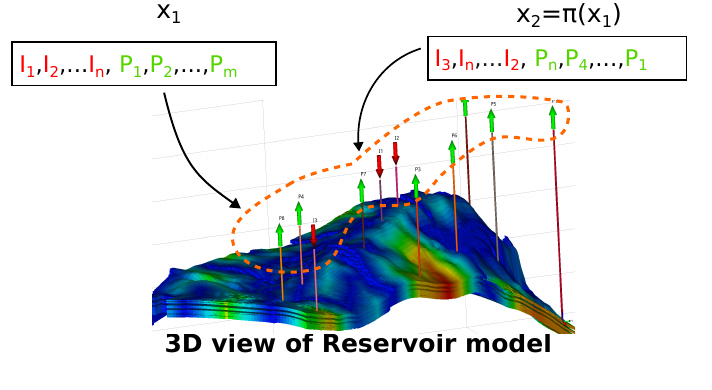}
\end{center}
\caption{Two vector encodings of the same physical well layout under a within-group permutation, $x_2=\pi(x_1)$. The simulator objective therefore satisfies $f(x_1)=f(x_2)$, whereas a kernel on flattened coordinates need not assign the pair self-similarity. This motivates enforcing permutation invariance in the surrogate. \label{fig1}}
\end{figure}

While powerful, standard BO surrogates like Gaussian Processes (GPs) \citep{williams2006gaussian} often assume a vector input space, which can be a critical limitation when the objective is invariant under a known group action and the surrogate does not encode this symmetry \citep{brown2024sample}. In many real-world problems, the input is not an unstructured vector but possesses inherent symmetries. For instance, in our CCS well-placement problem, the simulator output, and by extension the objective function, is invariant to the order in which injector well coordinates are listed within the injector group, and likewise invariant to permutations within the producer group. Equivalently, the objective is invariant under the action of the permutation group $S_m \times S_n$ acting separately on the $m$ injectors and $n$ producers. We claim that the Gaussian Process surrogate should exploit this information directly. A non-invariant GP treats permuted representations of the same physical design as distinct inputs, which can distort posterior geometry and lead to unnecessary exploration and reduced sample efficiency (Figure~\ref{fig1}).

Several approaches can encode permutation symmetries in BO surrogates. One direction is to engineer permutation-invariant set kernels, for example by aggregating pairwise element similarities (double-sum kernels) or by radializing the induced RKHS distance between sets, yielding analytic invariant GP kernels \citep{kim2021bayesian}. A second direction is to learn an invariant representation with a set encoder such as Deep Sets \citep{zaheer2017deep} and then place a GP on the learned embedding via Deep Kernel Learning (DKL) \citep{wilson2016deep}. The former offers an explicit inductive bias with no representation learning inside the BO loop, while the latter offers flexibility but introduces repeated nonconvex training from small, non-i.i.d.\ BO datasets.

We introduce \textbf{GP-Perm}, a permutation-invariant GP surrogate tailored to CCS well design under group control. GP-Perm enforces within-group exchangeability directly at the kernel level by comparing injector and producer configurations through a stable set discrepancy, while remaining compatible with standard vector-valued kernels for auxiliary continuous controls. Unlike pooled-similarity set kernels (e.g., double-sum / RKHS-distance constructions), GP-Perm is built around an explicit geometry-sensitive discrepancy between point sets. For completeness, we benchmark GP-Perm against (i) a learned-invariant baseline formed by combining DKL with a Deep Sets encoder (DKL-DS) \citep{wilson2016deep,zaheer2017deep}, (ii) engineered invariant set-kernel baselines (DS/DE) \citep{kim2021bayesian,buathong2020kernels}, and (iii) standard non-invariant GP/DKL surrogates. We evaluate these surrogates under a unified BO protocol across synthetic benchmarks (for controlled symmetry stress-tests) and simulation-based CCS tasks for realistic assessment.

The remainder of the paper is organized as follows: Section~\ref{problem_setup} formalizes the CCS design problem and its symmetry structure; Section~\ref{methodology} presents GP-Perm and DKL-DS; Sections~\ref{benchmarks}--\ref{ccs_cases} report benchmark and CCS results; and Section~\ref{sec:discussion} discusses implications and limitations.

\section{Background}
\label{background}

Bayesian optimization is well established for continuous vector spaces, but adapting its surrogate models to inputs with non-Euclidean or structured geometry is still an active area of research. Prominent examples include BO over strings \citep{moss2020boss} and combinatorial/graph-structured inputs \citep{oh2019combinatorial}. Our focus is on inputs that include unordered sets, where the objective is invariant to permutations of the set elements. This setting arises naturally in CCS well design under group control, where injector and producer well configurations form two unordered groups, accompanied by auxiliary continuous parameters. Equivalently, the objective is invariant under within-group permutations, i.e., under the action of $S_m \times S_n$ acting separately on injectors and producers. Accordingly, we develop an explicit kernel-based invariant GP surrogate (GP-Perm) and, for completeness, benchmark against a learned-invariant alternative implemented via Deep Sets within deep kernel learning (DKL-DS). These alternatives make different trade-offs in how faithfully they preserve geometry, how stable they are under the small, non-i.i.d.\ datasets encountered during BO, and how naturally they accommodate cross-set injector--producer interactions.

\paragraph{Engineered GP Kernels}
One approach to encoding set symmetry is to engineer permutation-invariant kernels. In CCS well placement, \citet{fotias2024optimization} solve the assignment problem with the Hungarian method to define a kernel distance. While effective, the resulting hard assignment is non-differentiable with respect to the inputs, which complicates gradient-based learning of kernel hyperparameters. More broadly, kernels based on hard matchings or raw distances can be brittle as small geometric perturbations may induce abrupt changes in correspondences or distances, and the induced notion of similarity can be difficult to tune beyond global lengthscale parameters. Related transport-based constructions have been explored for set inputs in other domains, e.g., an Earth Mover’s Distance (Wasserstein-1) kernel for sensor-set selection \citep{garnett2010bayesian} and an optimal-transport distance on neural architectures in NASBOT \citep{kandasamy2018neural}. This motivates using smooth transport-based discrepancies that retain geometric sensitivity while remaining amenable to stable hyperparameter learning inside BO. In contrast to using a raw set–set distance, the present work builds GP-Perm from a stable divergence between set representations that can be computed efficiently (e.g., via entropic transport solvers \citep{cuturi2013sinkhorn}), and composed with standard vector-valued kernels. This enables permutation-invariant BO for inputs containing multiple set components (injectors and producers) alongside auxiliary continuous variables.

\paragraph{Learned Permutation-Invariant Encoders within DKL}
A parallel line of work enforces invariance through neural architectures. Deep Sets \citep{zaheer2017deep} shows that any continuous permutation-invariant function on a set admits the form $\rho\left(\sum_{x\in S}\phi(x)\right)$, and the Set Transformer \citep{lee2019set} adds attention for richer inter-element interactions. For a recent overview of permutation-invariant neural networks and their variants (including how design choices such as the aggregation operator affect behavior), see \citep{kimura2024permutation}. Our DKL-DS model uses a Deep Sets encoder as a feature extractor within the Deep Kernel Learning (DKL) framework \citep{wilson2016deep}, which combines the representation power of deep neural networks with the probabilistic uncertainty quantification of GPs. In this paper, DKL-DS is included strictly as a learned invariant baseline against which we assess the benefits of encoding permutation invariance directly in the GP kernel. This baseline is especially relevant in expensive CCS BO, where the encoder is retrained repeatedly from small datasets and representation drift can alter posterior geometry and acquisition behavior across iterations.

\paragraph{Set and Distribution Kernels}
Several methods define kernels for non-vector inputs such as distributions or point clouds. A classical route is Maximum Mean Discrepancy (MMD) \citep{gretton2012kernel}, which embeds distributions into an RKHS. For set-valued inputs, RKHS-based set kernels provide permutation-invariant GPs: the Double-Sum kernel averages pairwise base-kernel similarities across elements and the Deep-Embedding (DE) kernel applies a radial kernel to the RKHS mean-embedding distance. DE is often strictly positive-definite and numerically stable \citep{buathong2020kernels}. Computationally efficient approximations make such set kernels practical for BO via unbiased subsampling and set-aware acquisition optimization \citep{kim2021bayesian}. However, these RKHS-based constructions often induce similarity through averaged pairwise comparisons or mean embeddings, which can smooth away fine geometric distinctions between layouts. In CCS design, where injector-producer placement depends on detailed spatial geometry and cross-set structure, this motivates complementary kernels that compare configurations through more geometry-sensitive discrepancies. Recent work also connects GP modeling on measures to entropic transport constructions \citep{bachoc2023gaussian}. In our experiments, we compare GP-Perm to both non-invariant surrogates and to these alternative set-kernel baselines.

\section{Methodology}
\label{methodology}
\subsection{Problem setup and Notation}
\label{problem_setup}
We consider Bayesian Optimization of an expensive black-box objective function
\begin{equation}
    \max_{\bm{x}\in\mathcal{X}} \; f(\bm{x}),
\end{equation}
where each evaluation of $f$ requires a high-fidelity reservoir simulation. The decision variable $\bm{x}$ has a structured form that reflects the physics and control of CCS well design. In particular, $\bm{x}$ contains (i) auxiliary continuous inputs that describe operational or contextual parameters and (ii) two unordered sets that describe the injector and producer well configurations.

We represent each design as
\begin{equation}
\bm{x} = \big( \bm{v}^{\bm{x}}, \; \mathcal{I}^{\bm{x}}, \; \mathcal{P}^{\bm{x}} \big),
\end{equation}
where $\bm{v}^{\bm{x}}\in\mathbb{R}^{d_v}$ denotes a vector of auxiliary continuous parameters (e.g., group control targets or other scalar settings). In our experiments, $\bm{v}^{\bm{x}}$ consists of the target rates of the injector and producer groups, so $d_v = 2$. Furthermore, $\mathcal{I}^{\bm{x}}=\{ \bm{i}_1,\dots,\bm{i}_{n_{\text{inj}}}\}$ is the set of injector well descriptors, and $\mathcal{P}^{\bm{x}}=\{ \bm{p}_1,\dots,\bm{p}_{n_{\text{prod}}}\}$ is the set of producer well descriptors. In this work, each well descriptor is a 2D spatial coordinate, i.e., $\bm{i}_j,\bm{p}_j\in\mathbb{R}^{2}$, but the formulation naturally extends to higher-dimensional well features (e.g., depth or completion parameters).

Under group control, the simulator output and the resulting objective depend on the injector and producer configurations only through the \emph{sets} $\mathcal{I}^{\bm{x}}$ and $\mathcal{P}^{\bm{x}}$, not on their internal ordering. Formally, for any permutations $\pi$ of $\{1,\dots,n_{\text{inj}}\}$ and $\sigma$ of $\{1,\dots,n_{\text{prod}}\}$,
\begin{equation}
f\!\left(\bm{v}^{\bm{x}}, \{\bm{i}_1,\dots,\bm{i}_{n_{\text{inj}}}\}, \{\bm{p}_1,\dots,\bm{p}_{n_{\text{prod}}}\}\right)
=
f\!\left(\bm{v}^{\bm{x}}, \{\bm{i}_{\pi(1)},\dots,\bm{i}_{\pi(n_{\text{inj}})}\}, \{\bm{p}_{\sigma(1)},\dots,\bm{p}_{\sigma(n_{\text{prod}})}\}\right).
\end{equation}
A surrogate model intended for BO should respect these symmetries to avoid redundant learning across equivalent permutations of the same physical design. We seek a probabilistic surrogate $p(f\,|\,\mathcal{D})$ that can be trained on observations $\mathcal{D}=\{(\bm{x}_t,y_t)\}_{t=1}^N$ with $y_t=f(\bm{x}_t)+\epsilon_t$, and that supports BO via an acquisition function. In the following subsections we construct (i) a permutation-invariant Gaussian Process surrogate by defining kernels that operate on the structured input $(\bm{v},\mathcal{I},\mathcal{P})$, and (ii) a learned invariant baseline using Deep Kernel Learning with Deep Sets.

\subsection{GP-Perm: a divergence-based permutation-invariant Mat\'ern kernel}
\label{gpperm}

To build a GP surrogate that is invariant to the ordering of wells, we define a kernel $k(\bm{x},\bm{x}')$ whose notion of similarity is itself permutation-invariant with respect to the injector and producer sets. Standard metrics such as Euclidean distance on the full input vector are sensitive to the ordering of set elements and are therefore unsuitable. Our approach is to construct a composite distance $D(\bm{x},\bm{x}')$ that respects the structured input $\bm{x} = \big(\bm{v}^{\bm{x}},\mathcal{I}^{\bm{x}},\mathcal{P}^{\bm{x}}\big)$ introduced in Section~\ref{problem_setup}.

\begin{figure}[t!]
	\begin{center}
		\includegraphics[width=.6\linewidth,keepaspectratio]{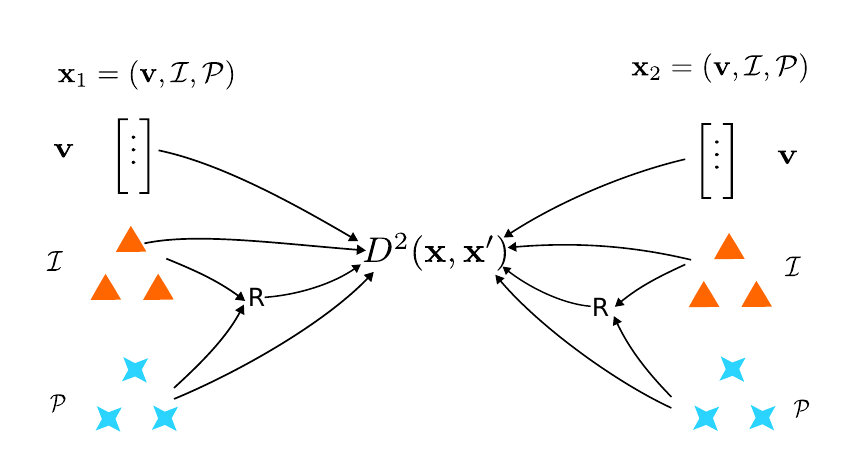}
	\end{center}
	\caption{GP-Perm kernel construction. The left and right panels show two inputs, each decomposed into an auxiliary vector component (e.g., group injection and production targets) and two unordered sets of well locations: injectors (orange) and producers (cyan). GP-Perm measures similarity by combining a conventional ARD distance on the auxiliary inputs with permutation-invariant Sinkhorn divergences between injector sets, producer sets, and the interaction sets $\bm{R}$ and $\bm{R}'$ defined in \eqref{eq:interaction_set}. These components form a single composite distance used by the GP; no ordering information within a set is used.
		\label{fig2}}
\end{figure}

We define a squared distance as an ANOVA-style sum of contributions from the auxiliary vector component, the injector and producer set geometries, and an  injector-producer interaction term (Figure~\ref{fig2}):

\begin{equation}
\label{eq:composite_distance_ip}
D^2(\bm{x},\bm{x}')
=
\sum_{j=1}^{d_v} \frac{\big(v_{j}-v'_{j}\big)^2}{\ell_{v,j}^2}
+
\frac{S_{\varepsilon}\!\big(\mathcal{I},\mathcal{I}'\big)}{\ell_{I}^2}
+
\frac{S_{\varepsilon}\!\big(\mathcal{P},\mathcal{P}'\big)}{\ell_{P}^2}
+
\frac{S_{\varepsilon}\!\big(\mathcal{R},\mathcal{R}'\big)}{\ell_{IP}^2},
\end{equation} 
where $\bm{v}\in\mathbb{R}^{d_v}$ denotes the vector-valued design variables, and $\mathcal{I}=\{\bm{i}_a\}_{a=1}^{n_{\mathrm{inj}}}\subset\mathbb{R}^2$ and $\mathcal{P}=\{\bm{p}_b\}_{b=1}^{n_{\mathrm{prod}}}\subset\mathbb{R}^2$ denote the injector and producer sets, respectively. The additional interaction set is defined as
\begin{equation}
\label{eq:interaction_set}
\mathcal{R}
=
\big\{\bm{r}_{ab}\big\}_{a=1,b=1}^{n_{\mathrm{inj}},n_{\mathrm{prod}}}
=
\big\{\bm{p}_b-\bm{i}_a \;:\; \bm{i}_a\in\mathcal{I},\;\bm{p}_b\in\mathcal{P}\big\}
\subset\mathbb{R}^2,
\end{equation}

This interaction set encodes cross-group spacing and relative positioning, which are central to CCS performance. Injectors and producers must be arranged to balance pressure management, plume front expansion and operational feasibility. The marginal geometries of $\bm{I}$ and $\bm{P}$ alone do not uniquely determine injector-producer spacing (e.g., two designs can share similar injector and producer layouts yet differ substantially in cross-group alignment), motivating the explicit inclusion of $S_{\varepsilon}(\mathcal{R},\mathcal{R}')$ in \eqref{eq:composite_distance_ip}.

In our experiments, $\bm{v}^{\bm{x}}$ consists of group injector and group producer control targets (so $d_v=2$), but the formulation directly supports higher-dimensional auxiliary inputs.

The set terms are computed using the \emph{Sinkhorn divergence} $S_{\varepsilon}$ between the injector sets and between the producer sets. Concretely, for two injector sets $\mathcal{I}=\{\bm{i}_a\}_{a=1}^{n_{\mathrm{inj}}}$ and
$\mathcal{I}'=\{\bm{i}'_a\}_{a=1}^{n_{\mathrm{inj}}}$ (and analogously for $\mathcal{P},\mathcal{P}'$ and $\mathcal{R},\mathcal{R}'$),
we form the pairwise cost matrix
\begin{equation}
C_{ab} = \|\bm{s}_a - \bm{t}_b\|_2,
\end{equation}
and compute the entropically regularized OT cost $W_{\varepsilon}(S,T)$ between the associated empirical measures with uniform weights, where $S=\{\bm{s}_a\}_{a=1}^{m}$ and $T=\{\bm{t}_b\}_{b=1}^{m}$ denote the two sets being compared (e.g., $S=\mathcal{I}$ and $T=\mathcal{I}'$). We then debias this quantity via
\begin{equation}
\label{eq:sinkhorn_divergence}
S_{\varepsilon}(S,T) \;=\; W_{\varepsilon}(S,T)
-\frac{1}{2}W_{\varepsilon}(S,S)
-\frac{1}{2}W_{\varepsilon}(T,T),
\end{equation}
which enforces $S_{\varepsilon}(S,S)=0$ and empirically improves numerical behavior compared to using $W_{\varepsilon}$ directly. In practice we clamp $S_{\varepsilon}$ to be non-negative to guard against small numerical violations. Since our application involves small sets of well locations, the cost of computing these set divergences is negligible compared to a single reservoir simulation. The entropic regularization parameter $\varepsilon$ controls the geometry of the discrepancy, interpolating between OT-like matching at small $\varepsilon$ and an MMD-like, pairwise-kernel regime at large $\varepsilon$, thereby linking GP-Perm to the behavior of classical double-sum set kernels in the high-regularization limit \cite{feydy2019interpolating}.

Finally, we construct GP-Perm by placing a Mat\'ern kernel on the composite distance:
\begin{equation}
k_{\text{GP-Perm}}(\bm{x},\bm{x}') = \sigma^2 \, \mathcal{M}_{\nu}\!\left(D(\bm{x},\bm{x}')\right),
\end{equation}
where $\nu$ controls smoothness and $\sigma^2$ is an output scale. Let $\pi\in S_{n_{\mathrm{inj}}}$ and $\tau\in S_{n_{\mathrm{prod}}}$, and define
$\pi(\mathcal{I})=\{\bm{i}_{\pi(a)}\}_{a=1}^{n_{\mathrm{inj}}}$ and
$\tau(\mathcal{P})=\{\bm{p}_{\tau(b)}\}_{b=1}^{n_{\mathrm{prod}}}$.
Since $S_{\varepsilon}$ depends only on the induced empirical measures, we have
$S_{\varepsilon}(\mathcal{I},\mathcal{I}')=S_{\varepsilon}(\pi(\mathcal{I}),\mathcal{I}')$ and
$S_{\varepsilon}(\mathcal{P},\mathcal{P}')=S_{\varepsilon}(\tau(\mathcal{P}),\mathcal{P}')$
(and likewise for $\mathcal{R}$).

Hence $D(\bm{x},\bm{x}')$ and therefore $k_{\text{GP-Perm}}(\bm{x},\bm{x}')$ are invariant under $S_{n_{\mathrm{inj}}}\times S_{n_{\mathrm{prod}}}$.

Optimizing hyperparameters of composite kernels that include set divergences can lead to ill-conditioned covariance matrices during marginal likelihood evaluation. We therefore adopt a staged jitter strategy: if Cholesky factorization fails at a nominal jitter level, we retry optimization under progressively larger diagonal jitter values. In our experiments this procedure was sufficient to obtain stable fits across all runs.

Even after successful fitting, posterior covariance computations can occasionally be sensitive to numerical roundoff when the kernel matrix is nearly singular. As a minimal safeguard, we clamp small negative eigenvalues that may arise from numerical error, ensuring a valid positive semidefinite covariance and hence a well-defined predictive distribution for acquisition optimization.
Appendix~\ref{appendixd} separates a conditional global PSD criterion from finite-design guarantees and relates these guarantees to the kernel matrices encountered in the experiments.

\begin{figure}[t!]
	\begin{center}
		\includegraphics[width=1.0\linewidth,keepaspectratio]{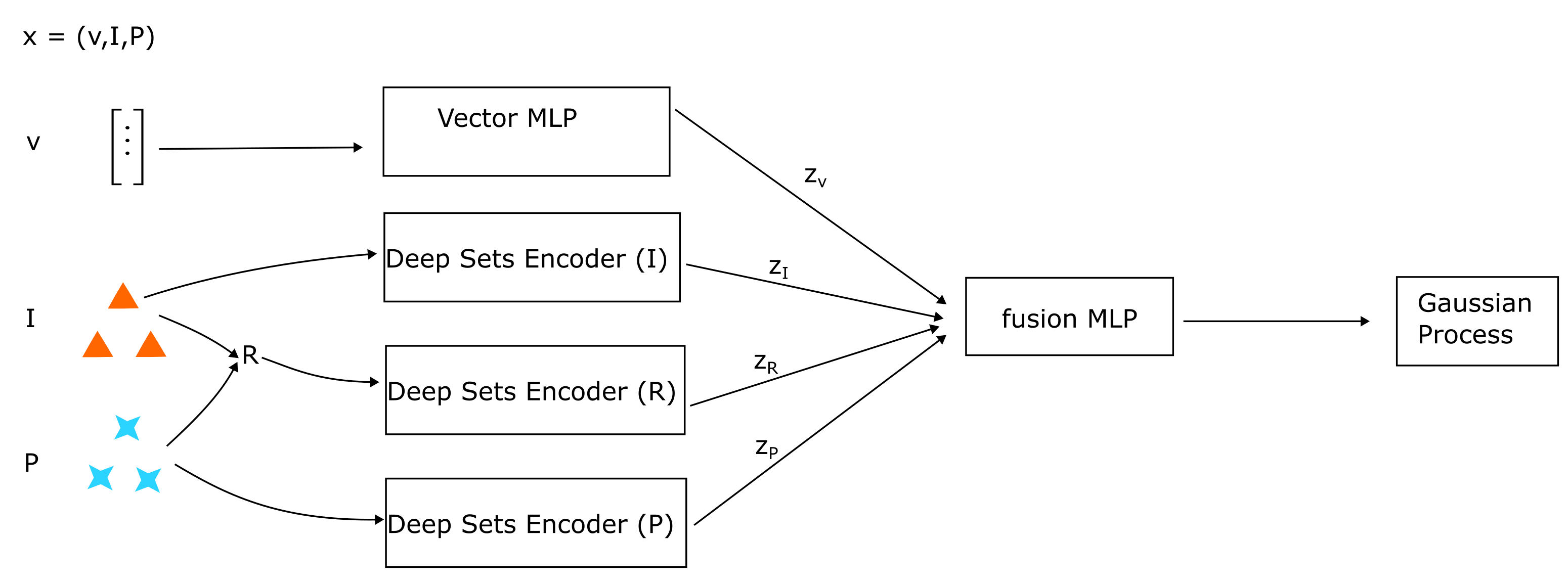}
	\end{center}
	\caption{DKL-DS architecture. An MLP embeds the auxiliary vector inputs, while Deep Sets encoders embed the injector, producer, and interaction sets using shared element-wise maps and invariant pooling; the fused latent representation is modeled by a GP in the latent space. \label{fig3}}
	
\end{figure}

\subsection{DKL-DS: a learned permutation-invariant surrogate baseline}
\label{dklds}

As a learned-invariant baseline to complement explicit kernel design, we consider a permutation-invariant surrogate based on Deep Kernel Learning (DKL) \citep{wilson2016deep}. In DKL, a neural feature extractor $f_{\theta}$ maps inputs to a latent space, and a GP is placed on the latent representation. Concretely, for $\bm{z}=f_{\theta}(\bm{x})$, the GP kernel operates as
\begin{equation}
k(\bm{x},\bm{x}') \;=\; k_{\text{lat}}\big(f_{\theta}(\bm{x}), f_{\theta}(\bm{x}')\big),
\end{equation}
where $k_{\text{lat}}$ is a standard kernel (e.g., Mat\'ern or RBF) defined on the latent space. This yields a GP model in a learned latent space, with uncertainty inherited from the GP posterior. Unlike GP-Perm, which enforces invariance directly through the kernel geometry, DKL-DS learns an invariant representation from data and places a standard GP on the resulting latent space.

To ensure invariance to permutations within injector and producer groups, we instantiate $f_{\theta}$ using Deep Sets \citep{zaheer2017deep}. Given the structured input $\bm{x}=\big(\bm{v}^{\bm{x}},\mathcal{I}^{\bm{x}},\mathcal{P}^{\bm{x}}\big)$ from Section~\ref{problem_setup}, our encoder processes the components separately and then fuses them (Figure~\ref{fig3}). In addition to $\mathcal{I}^{\bm{x}}$ and $\mathcal{P}^{\bm{x}}$, we construct the injector--producer interaction set $\mathcal{R}^{\bm{x}}$ as in \eqref{eq:interaction_set}, providing a permutation-invariant representation of cross-group spatial relationships.

\begin{enumerate}
    \item The auxiliary vector input $\bm{v}^{\bm{x}}$ is mapped through an MLP to a fixed-size embedding.
    \item The injector set $\mathcal{I}^{\bm{x}}$, producer set $\mathcal{P}^{\bm{x}}$, and interaction set $\mathcal{R}^{\bm{x}}$ are each processed by a permutation-invariant Deep Sets module.
    Each module first applies a shared element-wise map $\phi(\cdot)$ and then aggregates across the set
    using order-invariant pooling. In our implementation, we enrich the basic Deep Sets pooling by concatenating several invariant aggregations (mean, max, standard deviation, and power-means) before applying a final map $\rho(\cdot)$.
    \item The vector embedding and the three set embeddings are concatenated and passed through a fusion MLP to produce the final latent representation $\bm{z}=f_{\theta}(\bm{x})$.
    
\end{enumerate}
This construction makes $f_{\theta}$ permutation-invariant within each set by design, while the interaction set $\mathcal{R}^{\bm{x}}$ provides a dedicated channel for learning cross-group injector--producer geometric structure. The hybrid model is trained end-to-end by maximizing the GP marginal log-likelihood, allowing gradients to backpropagate through both the GP hyperparameters (e.g., latent-space lengthscales, optionally with ARD) and the network parameters $\theta$. We include DKL-DS to represent a strong AI engineered learned-invariance baseline, but note that it introduces repeated nonconvex training inside the BO loop, which can affect stability in the small-data regime. For improved optimization in the BO loop, we optionally warm-start $f_{\theta}$ using a short MSE pretraining phase with a linear prediction head, and then switch to joint DKL training. We also employ a staged jitter schedule to ensure robust Cholesky-based marginal likelihood evaluation.

\subsection{Bayesian Optimization loop}
\label{bo_loop}

We optimize the expensive black-box objective using a standard Bayesian Optimization loop. Starting from an initial dataset $\mathcal{D}_0=\{(\bm{x}_i, y_i)\}_{i=1}^{N_0}$, BO iterates between (i) fitting a probabilistic surrogate model to the accumulated data and (ii) selecting new candidate designs by maximizing an acquisition function.

We generate an initial design set by sampling the auxiliary continuous inputs uniformly within their bounds and sampling well locations from the feasible grid without replacement. When available, we bias the sampling towards interior feasible cells (\ref{appendixb}) using a pre-computed interior score, to reduce the likelihood of selecting wells close to invalid regions. Each initial design is evaluated with the reservoir simulator to obtain $y_i$.

At BO iteration $t$, we rescale each continuous input dimension to $[0,1]$ before fitting and acquisition optimization, which improves numerical conditioning for surrogate training and gradient-based acquisition maximization. We refit the surrogate on all available observations $\mathcal{D}_t$. If the simulator returns an invalid value (e.g., failed run) or a sentinel near-zero value indicating infeasibility, we assign it a constant penalty value chosen to be worse than the worst feasible observation observed so far. This ensures a well-defined training target while preserving the ordering among feasible outcomes.

We use the Monte Carlo acquisition function $q$LogEI \citep{ament2023unexpected} with a stochastic sampler. At each iteration, we select a batch of $q$ candidates by approximately maximizing $q$LogEI in the normalized continuous space using multi-start gradient-based optimization. Acquisition maximization is performed in the continuous, normalized space using multi-start gradient-based optimization. To promote feasible well placements during acquisition maximization, we augment the acquisition objective with a differentiable barrier penalty derived from a signed distance field over the feasible grid, with a user-controlled safety margin and penalty strength.

The optimized candidates are post-processed to produce valid discrete well locations. Specifically, each candidate well coordinate is snapped to a feasible grid cell by selecting among nearby feasible cells, favoring interior locations while avoiding duplicate wells within the same design. The resulting discrete designs are then unnormalized back to the original bounds and evaluated with the simulator. The new observations are appended to the dataset, $\mathcal{D}_{t+1}=\mathcal{D}_t \cup \{(\bm{x}_{t,j}, y_{t,j})\}_{j=1}^q$, and the process repeats for a fixed budget of BO iterations. The post-processing step enforces feasibility and uniqueness (no duplicate wells within a design) while remaining close to the optimized continuous candidate.

\subsection{Baseline surrogate models}
\label{baselines}

To contextualize the performance of GP-Perm and DKL-DS, we compare against four baselines that isolate the role of set-aware modeling:
\begin{itemize}
    \item \textbf{Non-invariant GP.}
    A standard Gaussian Process surrogate with a Mat\'ern kernel applied directly to the flattened input vector. This baseline does not encode permutation invariance and therefore treats different permutations of the same physical configuration as distinct inputs.

    \item \textbf{Non-invariant DKL.}
    A Deep Kernel Learning surrogate in which an MLP feature extractor maps the flattened input to a latent representation and a GP with a Mat\'ern kernel is placed on the learned features. Unlike DKL-DS, the encoder is not permutation-invariant.

    \item \textbf{Double-Sum (DS) set-kernel GP.}
    A classical permutation-invariant set kernel that aggregates a base kernel across all cross-element pairs between two sets (equivalently, the average pairwise similarity). This provides a strong set-aware GP baseline and is commonly used in set-valued BO comparisons \citep{buathong2020kernels,kim2021bayesian}. For grouped inputs, we apply DS separately to injectors and producers and combine the resulting terms additively with the auxiliary-vector kernel.

    \item \textbf{Deep-Embedding (DE) set-kernel GP.}
    A permutation-invariant set kernel based on RKHS mean embeddings: each set is mapped to its empirical kernel mean embedding under a chosen base kernel, and set similarity is computed by applying a radial kernel to the induced RKHS distance between embeddings. In this work, DE uses a fixed kernel mean embedding (not a learned neural embedding), serving as a stable and widely used set-kernel baseline \citep{buathong2020kernels,kim2021bayesian}. As above, DE is applied per set component and combined with the auxiliary-vector kernel using the same composite structure across experiments.

\end{itemize}

For inputs with multiple components (auxiliary vectors plus injector and producer sets), DS and DE are applied independently to the injector and producer sets and combined with a conventional kernel on the auxiliary inputs using the same additive composition used in \eqref{eq:composite_distance_ip}. All baselines are trained by maximizing the marginal log-likelihood and evaluated under the BO protocol in Section~\ref{bo_loop}.

\section{Benchmark study for model selection}
\label{benchmarks}
\begin{table}[t!]
\centering
\caption{Synthetic benchmarks and BO protocol used for model selection. All benchmarks operate on normalized coordinates in $[0,1]^2$. Here $n_{\text{points}}$ is the set cardinality and $\text{Dim}=2\,n_{\text{points}}$ is the flattened decision-space dimension.}
\label{tab:bench_protocol}
\begin{tabular}{@{}lcccc@{}}
\toprule
\textbf{Benchmark} &
$\mathbf{n_{\text{points}}}$ &
$\mathbf{\text{Dim}}$ &
$\mathbf{n_{\text{trials}}}$ &
$\mathbf{T}$ \\
\midrule
Particle Physics & $10$ & $20$ & $10$ & $20$ \\
Max Area Coverage & $16$ & $32$ & $10$ & $20$ \\
Distribution Matching (MMD) & $20$ & $40$ & $10$ & $10$ \\
Max Spanning Tree & $15$ & $30$ & $10$ & $10$ \\
Facility Location (Soft Coverage) & $12$ & $24$ & $8$ & $10$ \\
Soft $k$-Medoids & $12$ & $24$ & $10$ & $10$ \\
\bottomrule
\end{tabular}%
\end{table}
Before running expensive CCS case studies, we perform a benchmark-driven model selection study on a suite of synthetic permutation-invariant objectives. This stage serves two purposes: (i) to verify under controlled conditions that the proposed surrogates exploit permutation symmetry, and (ii) to select stable model variants and hyperparameters that are carried forward unchanged to the CCS case studies in Section \ref{ccs_cases}. Full definitions of the benchmark objectives are provided in Appendix \ref{appendixa}.

\subsection{Synthetic benchmark suite}
\label{synthetic_suite}

The synthetic benchmarks are constructed to be permutation-invariant by design and to stress different aspects of set reasoning (geometric structure, coverage, distributional matching, and graph-based objectives). Each benchmark takes as input a single unordered set of $n_{\text{points}}$ two-dimensional locations, represented in code as a flattened vector in $\mathbb{R}^{2n_{\text{points}}}$ but interpreted by the models as a set. Importantly, unlike the CCS setting, these benchmarks do not include auxiliary continuous inputs and do not include separate injector and producer sets.

Accordingly, during the benchmark stage we use reduced versions of our surrogate families that operate on a single set input. This isolates the effect of permutation-invariant modeling from CCS-specific components (auxiliary controls and multiple well groups) and allows us to compare kernel/encoder variants in a controlled regime. The selected configurations are then transferred to the CCS input structure (auxiliary vectors + two sets) without additional tuning.

\subsection{BO protocol and evaluation metrics}
\label{benchmark_protocol}

For each synthetic benchmark objective, we run BO for a fixed evaluation budget with multiple independent trials. All benchmarks follow the BO loop described in Section~\ref{bo_loop}, specialized to a single-set input: each trial begins with an initial design of size $N_{\text{init}}$, followed by $T$ BO iterations with batch size $q$. At each iteration, the surrogate is refit on all collected data and the acquisition function is optimized to propose the next batch of candidate sets. Benchmark-specific budgets are used to keep overall difficulty comparable across tasks.

We evaluate performance using best-so-far trajectories and two scalar summaries computed per trial: (i) the area under the best-so-far curve (AUC) and (ii) the final best value. Model selection within each family (GP-Perm and DKL-DS) is performed via a sensitivity study over hyperparameters; the full selection procedure and any score normalization used for variant selection are reported in ~\ref{appendixe}. Table~\ref{tab:bench_protocol} summarizes the benchmark settings used throughout this section.

\subsection{Benchmark comparison of selected models against baselines}
\label{benchmark_selection}

Beyond model selection, the synthetic benchmark suite provides a controlled setting to compare permutation-invariant surrogates against strong non-invariant and set-kernel baselines under identical BO budgets and protocols (Section~\ref{benchmark_protocol}). We first perform a hyperparameter ablation for GP-Perm and DKL-DS and select a single configuration of each based on benchmark performance; the complete sensitivity analysis and all variant results are reported in ~\ref{appendixe}. The two selected models (denoted \textbf{GP-Perm} and \textbf{DKL-DS} below) are then compared against (i) non-invariant baselines (\textbf{GP}, \textbf{DKL}) and (ii) set-kernel baselines (\textbf{DS-GP}, \textbf{DE-GP}), all evaluated with the same BO setup.

\begin{figure}[t]
    \centering
    \includegraphics[width=\linewidth]{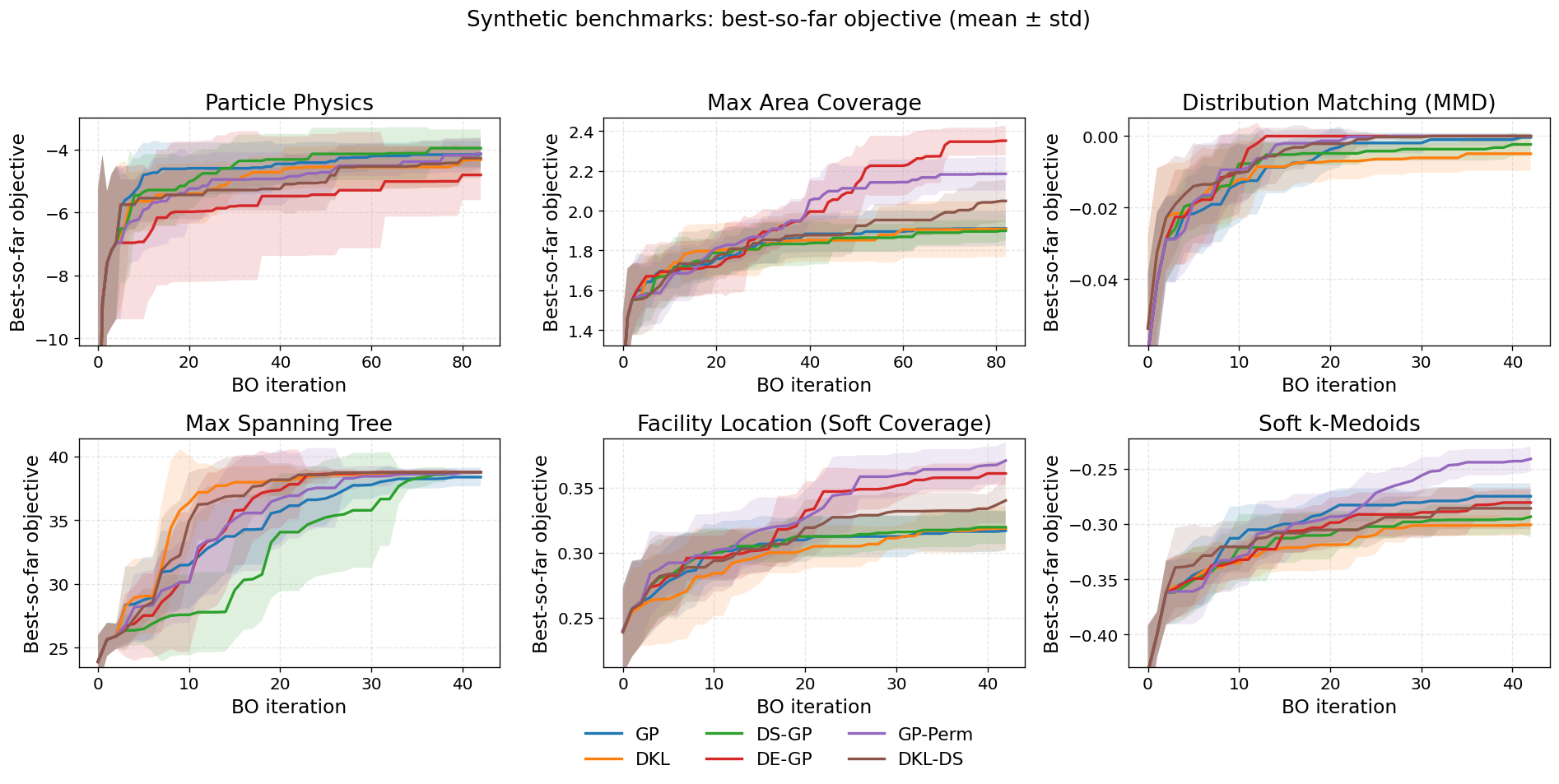}
    \caption{Synthetic benchmarks: best-so-far objective versus BO iteration (mean $\pm$ std across trials).}
    \label{fig:benchmark_plots}
\end{figure}

Figure~\ref{fig:benchmark_plots} shows the mean best-so-far objective across trials (shaded: $\pm 1$ std). Tables~\ref{tab:bench_auc}--\ref{tab:bench_final} summarize the two scalar metrics used throughout the study: the AUC of the best-so-far curve and the final best value, reported as mean $\pm$ std across trials (best per benchmark in bold). All objectives are maximized; for benchmarks whose range is negative, values closer to zero correspond to better performance. The selected GP-Perm and DKL-DS configurations are used \emph{unchanged} in the CCS case studies in Section~\ref{ccs_cases}.

\begin{table}[t]
\centering
\small
\setlength{\tabcolsep}{4pt}
\caption{Benchmark summary using AUC of the best-so-far curve (mean $\pm$ std across trials).}
\label{tab:bench_auc}
\resizebox{\linewidth}{!}{%
\begin{tabular}{lcccccc}
\toprule
\textbf{Benchmark} & \textbf{GP} & \textbf{DKL} & \textbf{DS-GP} & \textbf{DE-GP} & \textbf{GP-Perm} & \textbf{DKL-DS} \\
\midrule
Distribution Matching (MMD)          & -0.371 $\pm$ 0.096 & -0.425 $\pm$ 0.242 & -0.369 $\pm$ 0.112 & \textbf{-0.238 $\pm$ 0.104} & -0.272 $\pm$ 0.126 & -0.243 $\pm$ 0.141 \\
Facility Location (Soft Coverage)    & 12.720 $\pm$ 0.712 & 12.478 $\pm$ 0.703 & 12.806 $\pm$ 0.547 & 13.607 $\pm$ 0.522 & \textbf{13.826 $\pm$ 0.635} & 13.071 $\pm$ 0.512 \\
Max Area Coverage                    & 149.450 $\pm$ 5.887 & 149.373 $\pm$ 7.542 & 148.171 $\pm$ 4.986 & \textbf{163.894 $\pm$ 5.954} & 160.621 $\pm$ 5.660 & 151.845 $\pm$ 4.758 \\
Max Spanning Tree                    & 1452.410 $\pm$ 62.506 & \textbf{1524.105 $\pm$ 39.065} & 1368.808 $\pm$ 87.785 & 1474.793 $\pm$ 65.525 & 1465.257 $\pm$ 76.264 & 1507.203 $\pm$ 34.486 \\
Particle Physics                     & \textbf{-393.629 $\pm$ 48.427} & -425.810 $\pm$ 52.625 & -395.263 $\pm$ 64.242 & -482.277 $\pm$ 148.410 & -427.180 $\pm$ 48.192 & -434.479 $\pm$ 54.414 \\
Soft k-Medoids                       & -12.663 $\pm$ 0.469 & -13.509 $\pm$ 0.425 & -13.298 $\pm$ 0.446 & -13.057 $\pm$ 0.493 & \textbf{-12.365 $\pm$ 0.515} & -12.990 $\pm$ 0.591 \\
\bottomrule
\end{tabular}%
}
\end{table}

\begin{table}[!t]
\centering
\small
\setlength{\tabcolsep}{4pt}
\caption{Benchmark summary using final best value at the end of BO (mean $\pm$ std across trials; higher is better).}
\label{tab:bench_final}
\resizebox{\linewidth}{!}{%
\begin{tabular}{lcccccc}
\toprule
\textbf{Benchmark} & \textbf{GP} & \textbf{DKL} & \textbf{DS-GP} & \textbf{DE-GP} & \textbf{GP-Perm} & \textbf{DKL-DS} \\
\midrule
Distribution Matching (MMD)          & -0.000 $\pm$ 0.001 & -0.005 $\pm$ 0.005 & -0.002 $\pm$ 0.003 & \textbf{0.000 $\pm$ 0.000} &\textbf{0.000 $\pm$ 0.000} &\textbf{0.000 $\pm$ 0.000} \\
Facility Location (Soft Coverage)    & 0.317 $\pm$ 0.015 & 0.319 $\pm$ 0.017 & 0.320 $\pm$ 0.013 & 0.361 $\pm$ 0.009 & \textbf{0.371 $\pm$ 0.014} & 0.340 $\pm$ 0.018 \\
Max Area Coverage                    & 1.911 $\pm$ 0.088 & 1.908 $\pm$ 0.141 & 1.899 $\pm$ 0.051 & \textbf{2.352 $\pm$ 0.074} & 2.185 $\pm$ 0.084 & 2.049 $\pm$ 0.105 \\
Max Spanning Tree                    & 38.384 $\pm$ 0.722 & 38.734 $\pm$ 0.073 & 38.770 $\pm$ 0.000 & \textbf{38.772 $\pm$ 0.009} & 38.751 $\pm$ 0.042 & 38.769 $\pm$ 0.002 \\
Particle Physics                     & -4.153 $\pm$ 0.417 & -4.324 $\pm$ 0.451 & \textbf{-3.957 $\pm$ 0.594} & -4.807 $\pm$ 0.802 & -4.133 $\pm$ 0.510 & -4.274 $\pm$ 0.640 \\
Soft k-Medoids                       & -0.275 $\pm$ 0.012 & -0.301 $\pm$ 0.009 & -0.294 $\pm$ 0.019 & -0.281 $\pm$ 0.012 & \textbf{-0.241 $\pm$ 0.012} & -0.286 $\pm$ 0.019 \\
\bottomrule
\end{tabular}%
}
\end{table}
Overall, no single surrogate dominates every benchmark. DE-GP is strongest on Distribution Matching and Max Area Coverage, while DKL attains the best AUC on Max Spanning Tree; for Facility Location and Soft $k$-Medoids, the selected GP-Perm achieves the best performance under both AUC and final best. These results highlight that the benefit of permutation invariance is task dependent: enforcing the correct symmetry can yield clear gains, but the best-performing configuration within a model family can vary across objectives. In particular, the full sensitivity results in ~\ref{appendixe} show that some GP-Perm and DKL-DS variants attain the highest reported sample means on specific benchmarks even when the single configuration selected for transfer to CCS is not the per-benchmark winner. We therefore interpret the synthetic suite both as (i) evidence that permutation-invariant surrogates can be competitive or superior under controlled symmetries, and (ii) motivation for careful hyperparameter selection when targeting a specific task.

\section{CCS case study}
\label{ccs_cases}

\begin{table}[b!]
\centering
\caption{Surrogate configurations carried forward to CCS after benchmark selection.}
\label{tab:ccs_models}
\resizebox{\textwidth}{!}{%
\begin{tabular}{@{}lp{0.82\textwidth}@{}}
\toprule
\textbf{Model} & \textbf{Key settings} \\
\midrule
GP (non-invariant) &
kernel:matern52. \\

DKL (non-invariant) &
latent\_dim:64; mlp\_hidden:64; kernel:matern52; ARD:true. \\

GP-DS (set-kernel baseline) &
vec\_kernel:matern52; set\_kernel:DS (RBF base) applied to $\mathcal{I}^{\bm{x}}$, $\mathcal{P}^{\bm{x}}$, and $\mathbb{R}^{\bm{x}}$. \\

GP-DE (set-kernel baseline) &
vec\_kernel:matern52; set\_kernel:DE (RBF base + RBF outer) applied to $\mathcal{I}^{\bm{x}}$, $\mathcal{P}^{\bm{x}}$, and $\mathbb{R}^{\bm{x}}$. \\

GP-Perm (selected) &
kernel:matern52; $\varepsilon$: benchmark-selected (Table~\ref{tab:variant_sweeps}). \\

DKL-DS (selected) &
latent\_dim:64; $\phi$:64; $\rho$:128; pools: benchmark-selected (Table~\ref{tab:variant_sweeps}); kernel:matern52; +ARD if selected; pretrain\_epochs:100. \\
\bottomrule
\end{tabular}%
}
\end{table}

\subsection{CCS-like two-set synthetic benchmark}
\label{ccs_synth}

Before turning to expensive reservoir simulation, we validate our permutation-invariant surrogate on a controlled two-set benchmark that captures key geometric features of CCS well design under grouped injectors and producers. The input is composed of two unordered sets, injector and producer locations and the objective is constructed so that each producer is encouraged to have (at least) one injector at a preferred spacing and along a preferred migration direction, while within-set repulsion discourages collapse of injector--injector and producer--producer locations.

Concretely, for $n_{\text{inj}}$ injectors and $n_{\text{prod}}$ producers in a bounded 2D domain, a candidate design is a flattened vector of coordinates
$
\bm{x}=\big[(\bm{i}_1,\dots,\bm{i}_{n_{\text{inj}}}),(\bm{p}_1,\dots,\bm{p}_{n_{\text{prod}}})\big],
$
with $\bm{i}_k,\bm{p}_\ell\in[-1,1]^2$. For each producer $\bm{p}_\ell$, we define a pairwise mismatch cost to each injector $\bm{i}_k$ that penalizes deviation from a preferred distance $d^\star$ and misalignment with a preferred direction $\bm{u}$ (normalized):
\begin{equation}
c_{\ell k}
=
\left(\frac{\|\bm{p}_\ell-\bm{i}_k\|-d^\star}{\sigma_d}\right)^2
+
\left(\frac{1-\cos(\angle(\bm{p}_\ell-\bm{i}_k),\bm{u})}{\sigma_\theta}\right)^2,
\end{equation}
and we aggregate over injectors via a soft minimum (temperature $\tau$) to reflect that each producer mainly depends on its most effective injector:
\begin{equation}
C_{\text{cross}}(\bm{x})
=
\frac{1}{n_{\text{prod}}}\sum_{\ell=1}^{n_{\text{prod}}}
\Big(
-\tau\log\sum_{k=1}^{n_{\text{inj}}}\exp(-c_{\ell k}/\tau)
\Big).
\end{equation}
Finally, we add within-set repulsion terms for injectors and producers (inverse squared distances with small $\varepsilon$) to discourage degenerate clustering:
\begin{equation}
C_{\text{rep}}(\bm{x})
=
w_{\text{inj}}\sum_{k<k'}\frac{1}{\|\bm{i}_k-\bm{i}_{k'}\|^2+\varepsilon}
+
w_{\text{prod}}\sum_{\ell<\ell'}\frac{1}{\|\bm{p}_\ell-\bm{p}_{\ell'}\|^2+\varepsilon}.
\end{equation}
The benchmark returns a scalar objective $f(\bm{x})=-\big(C_{\text{cross}}(\bm{x})+C_{\text{rep}}(\bm{x})\big)$ (higher is better), and is permutation-invariant within the injector set and within the producer set by construction. This mirrors the symmetry induced by grouped well control in CCS, where only the unordered configuration of injectors and producers matters.

\subsection{Reservoir simulator case study: Johansen formation}
\label{ccs_setup}

We next evaluate the same surrogate models on a realistic CCS optimization task derived from one of the only large-scale, publicly available geological models for prospective CO$_2$ storage: the Johansen formation \citep{andersen2018co2}. We use the structured design variable from Section~\ref{problem_setup}, i.e.,
$\bm{x}=\big(\bm{v}^{\bm{x}},\mathcal{I}^{\bm{x}},\mathcal{P}^{\bm{x}}\big)$,
where $\bm{v}^{\bm{x}}$ denotes group control targets and
$\mathcal{I}^{\bm{x}},\mathcal{P}^{\bm{x}}$ are unordered well-location sets.

A candidate design consists of (i) the $(x,y)$ locations of each injector and producer well and (ii) auxiliary continuous operational controls given by group target rates (injection and production). The objective is to maximize a regularized Net Present Value (NPV) that incorporates penalties for undesirable operational outcomes such as CO$_2$ recycling and rate inconsistency. Each evaluation requires a computationally expensive run of the OPM-Flow reservoir simulator \citep{rasmussen2021open}. In addition to the active injection/production period, each simulation includes a post-operation monitoring period of 50 years:
\begin{equation}
\label{eq:ccs_objective}
f(\bm{x}) \;=\; \text{NPV}\bigl(\text{Sim}(\bm{x})\bigr) \;-\; P_L\bigl(\text{Sim}(\bm{x})\bigr) \;-\; P_S\bigl(\text{Sim}(\bm{x})\bigr).
\end{equation}

\subsection{Feasible well-placement regions and candidate generation}
\label{ccs_feasible}

A key challenge in CCS well placement is defining a realistic search space. Valid well locations form a discrete, non-convex region constrained by the reservoir geometry and operational restrictions. For Johansen, we exclude the thin western part of the formation to mitigate potential leakage risks \citep{bergmo2009exploring,bergmoa2009exploring}. Visualizations of the feasible region before and after masking are provided in Appendix~\ref{appendixb}.

The resulting mask defines the feasible grid cells used during Bayesian optimization. Candidate generation and feasibility handling follow the procedure described in Subsection~\ref{bo_loop}: acquisition optimization is performed in the normalized continuous design space, and proposed well coordinates are post-processed onto valid grid cells before simulator evaluation. This ensures that all evaluated designs satisfy the geological feasibility constraints while keeping the acquisition-optimization step compatible with gradient-based BO.

\subsection{Evaluation protocol}
\label{ccs_cases_def}

\textbf{Synthetic benchmark.}
We evaluate the models in Table~\ref{tab:ccs_models} on the CCS-like two-set synthetic function with
$n_{\text{inj}}=3$, $n_{\text{prod}}=5$,
$n_{\text{trials}}=9$ independent trials,
$N_{\text{init}}=8$ initial Sobol designs,
$T=30$ BO iterations,
and batch size $q=4$ per iteration.
We report mean best-so-far objective values across trials, with variability bands given by mean $\pm$ one standard deviation.

\textbf{Johansen case study.}
We then evaluate the same surrogate models on the Johansen formation under a tighter evaluation budget due to the cost of OPM-Flow simulations:
$n_{\text{trials}} = 7$ independent trials,
$N_{\text{init}} = 3$ initial Sobol designs,
$T = 10$ BO iterations,
and batch size $q = 2$ per iteration.
In preliminary experiments we observed persistent simulator instability in the Smeaheia setup under repeated BO evaluations, preventing reliable trial-based comparisons under the same protocol; we therefore focus on Johansen for the realistic CCS study, while retaining feasibility construction and visualizations for both formations in Appendix \ref{appendixb}.

\subsection{Results: CCS-like synthetic and Johansen}
\label{ccs_results}

\begin{figure}[t!]
	\centering
	\begin{subfigure}[t]{0.49\textwidth}
		\centering
		\includegraphics[width=\textwidth]{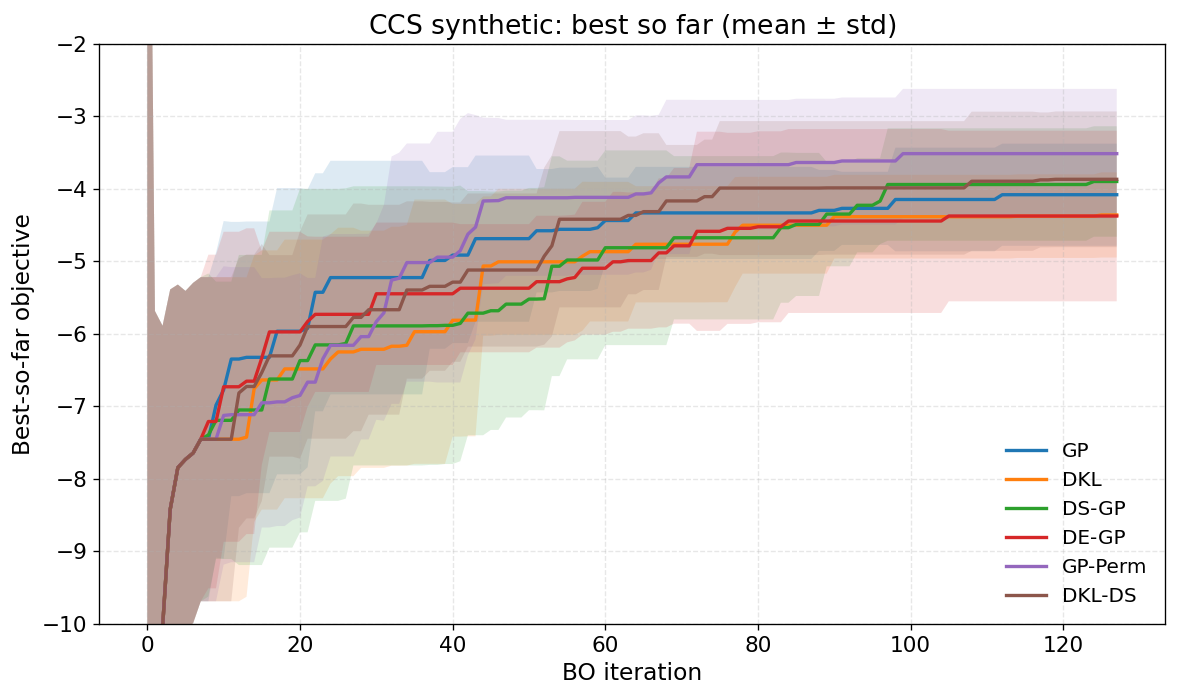}
		\caption{CCS-like two-set synthetic benchmark.}
		\label{fig:ccs_synth_plot}
	\end{subfigure}\hfill
	\begin{subfigure}[t]{0.49\textwidth}
		\centering
		\includegraphics[width=\textwidth]{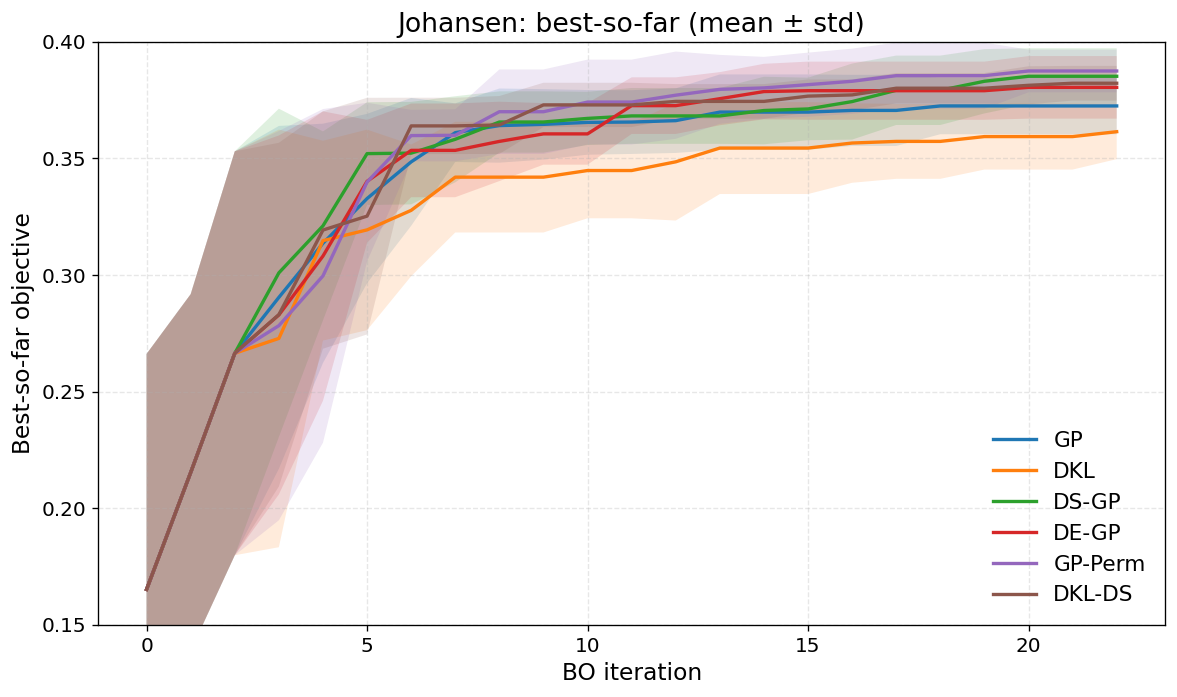}
		\caption{Johansen CCS case study.}
		\label{fig:johansen_plot}
	\end{subfigure}
	\caption{BO results on (left) CCS-like two-set synthetic benchmark and (right) Johansen CCS case study: mean best-so-far trajectories (shaded: $\pm 1$ std).}
	\label{fig:ccs_two_panel}
\end{figure}

Figure~\ref{fig:ccs_two_panel} reports best-so-far trajectories for the CCS-like two-set synthetic benchmark (left) and the Johansen CCS case study (right). In both settings, GP-Perm achieves the highest mean AUC and mean final objective among the tested models. These rankings of sample means are descriptive and do not imply statistical separation between methods.

\begin{table}[!t]
	\centering
	\caption{AUC of best-so-far curves (mean $\pm$ std across trials) for the CCS-like synthetic benchmark and the Johansen case study.}
	\label{tab:ccs_auc_combined}
	\begin{tabular}{lcc}
		\toprule
		Model & Synthetic AUC & Johansen AUC \\
		\midrule
		GP & -633.905 $\pm$ 99.069 & 7.563 $\pm$ 0.467 \\
		DKL & -688.067 $\pm$ 113.562 & 7.256 $\pm$ 0.486 \\
		DS-GP & -674.499 $\pm$ 124.006 & 7.672 $\pm$ 0.336 \\
		DE-GP & -673.955 $\pm$ 115.208 & 7.626 $\pm$ 0.267 \\
		GP-Perm & \textbf{-603.991 $\pm$ 120.954} & \textbf{7.717 $\pm$ 0.479} \\
		DKL-DS & -636.862 $\pm$ 89.690 & 7.675 $\pm$ 0.277 \\
		\bottomrule
	\end{tabular}
\end{table}

\begin{table}[!t]
	\centering
	\caption{Final best-so-far objective value (mean $\pm$ std across trials) for the CCS-like synthetic benchmark and the Johansen case study.}
	\label{tab:ccs_final_combined}
	\begin{tabular}{lcc}
		\toprule
		Model & Synthetic FINAL & Johansen FINAL \\
		\midrule
		GP & -4.082 $\pm$ 0.704 & 0.372 $\pm$ 0.012 \\
		DKL & -4.362 $\pm$ 0.587 & 0.361 $\pm$ 0.012 \\
		DS-GP & -3.900 $\pm$ 0.762 & 0.385 $\pm$ 0.012 \\
		DE-GP & -4.376 $\pm$ 1.177 & 0.380 $\pm$ 0.013 \\
		GP-Perm & \textbf{-3.515 $\pm$ 0.893} & \textbf{0.387 $\pm$ 0.009} \\
		DKL-DS & -3.869 $\pm$ 0.936 & 0.382 $\pm$ 0.007 \\
		\bottomrule
	\end{tabular}
\end{table}

On the CCS-like synthetic benchmark, GP-Perm attains the highest mean AUC (Table~\ref{tab:ccs_auc_combined}), increasing from $-633.905$ for GP to $-603.991$, and the highest mean final objective (Table~\ref{tab:ccs_final_combined}), increasing from $-4.082$ for GP to $-3.515$. Its corresponding sample means also exceed those of DS-GP and DE-GP. This setting isolates the core structure targeted by GP-Perm: the objective depends on relative geometry between two unordered sets while being insensitive to within-set permutations. Under this fixed protocol, GP-Perm therefore has the strongest mean best-so-far summaries, with the across-trial dispersion shown in Tables~\ref{tab:ccs_auc_combined}--\ref{tab:ccs_final_combined}.

The same ranking of sample means is observed for the realistic simulator-based CCS objective on Johansen, despite a substantially tighter BO budget ($T{=}10$, $q{=}2$) than in the synthetic study ($T{=}30$, $q{=}4$). GP-Perm achieves the highest mean AUC (7.717) and mean final objective (0.387) among the tested models (Tables~\ref{tab:ccs_auc_combined}--\ref{tab:ccs_final_combined}). These results are consistent with a useful permutation-invariant inductive bias in both the CCS-like setting and the high-fidelity simulator, but the reported variability does not establish pairwise statistical superiority.

\section{Discussion}
\label{sec:discussion}

This work studies Bayesian optimization for CCS well design when the input combines unordered well-location sets with auxiliary continuous controls. Under group control, the objective is invariant to within-group permutations of wells, so non-invariant surrogates waste capacity modeling distinctions that are physically meaningless and can distort posterior geometry under tight evaluation budgets. Our primary contribution is GP-Perm: a permutation-invariant Gaussian process surrogate that enforces this symmetry directly through an explicit set-based kernel built from a stable divergence between well configurations. For completeness, we also evaluate a learned-invariant baseline, DKL-DS, which implements the same symmetry indirectly by training a Deep Sets encoder and placing a standard GP in the induced latent space. Finally, we include non-invariant GP/DKL and set-kernel baselines (DS/DE) to isolate the effect of explicit invariance versus pooled representations. Model configurations are selected via a benchmark-driven stage and then transferred unchanged to the CCS setting. Simulator-free diagnostics in Appendix~\ref{appendixd} independently verify permutation invariance to numerical precision, characterize the kernel's response to geometric perturbations and regularization, and show that the interaction block distinguishes changes in cross-group geometry that are invisible to the marginal injector and producer comparisons.

Beyond CCS, our results support a general lesson for sequential, low-data learning: when inputs contain nuisance symmetries, encoding the symmetry reduces effective complexity by collapsing equivalent designs. In BO, this can translate into a smoother posterior over the quotient space of designs and consequently, more reliable acquisition decisions under small evaluation budgets. In our synthetic suite, permutation-invariant surrogates are frequently competitive with or superior to non-invariant GP/DKL baselines under identical BO budgets, although the benchmark tables contain exceptions and no model family wins throughout. This is expected because the benchmarks stress different aspects of set reasoning (coverage, geometric structure, distributional matching, and graph-like interactions), and each surrogate family imposes a different inductive bias. To further connect the abstract benchmarks to CCS structure, we also introduced a CCS-like two-set synthetic objective that mimics injector--producer geometric coupling under group control while preserving within-group exchangeability. This benchmark provides a controlled setting in which the symmetry targeted by GP-Perm is exact, while still reflecting the interaction-driven nature of CCS well layouts. Because BO performance depends not only on final objective values but also on posterior conditioning and acquisition stability, we report additional diagnostics that summarize how each surrogate shapes uncertainty and acquisition behavior across iterations on Johansen.

\begin{table}[t!]
	\centering
	\caption{Johansen posterior and acquisition diagnostics (mean $\pm$ std across trials; averaged over BO iterations 1--9). Here $\rho_{\mathrm{qLogEI}}=\mathrm{qLogEI}(\bm{x}_{\text{next}})/\max_{\bm{z}\in Z}\mathrm{qLogEI}(\bm{z})$ is the raw quotient used in the diagnostic script relative to a fixed Sobol probe set $Z$. Since qLogEI is log-scaled, $\rho_{\mathrm{qLogEI}}$ should be interpreted only as an implementation-specific relative diagnostic, not as a calibrated EI-scale acquisition ratio. $\mathbb{E}_{Z}[\sigma(\bm{x})]$ is the mean posterior standard deviation over $Z$, $\sigma(\bm{x}_{\text{next}})$ is the posterior standard deviation at the selected batch, and $|\bar r|$ is the absolute value of the batch-averaged standardized residual at the sampled points.}
	\label{tab:johansen_diagnostics}
	\begin{tabular}{lcccc}
		\toprule
		Model & $\rho_{\mathrm{qLogEI}}$ & $\mathbb{E}_{Z}[\sigma(\bm{x})]$ & $\sigma(\bm{x}_{\text{next}})$ & $|\bar r|$ \\
		\midrule
		GP      & 1 $\,\pm\,$ 0.0369 & 0.0776 $\,\pm\,$ 0.0138 & 0.0626 $\,\pm\,$ 0.0120 & 0.459 $\,\pm\,$ 0.140 \\
		DKL     & 3.22 $\,\pm\,$ 2.72 & 0.00291 $\,\pm\,$ 0.000447 & 0.00418 $\,\pm\,$ 0.00124 & 17.7 $\,\pm\,$ 4.09 \\
		DS-GP   & 1.03 $\,\pm\,$ 0.298 & 0.0193 $\,\pm\,$ 0.0197 & 0.0182 $\,\pm\,$ 0.0196 & 1.63 $\,\pm\,$ 0.432 \\
		DE-GP   & 1.51 $\,\pm\,$ 0.601 & 0.0142 $\,\pm\,$ 0.00985 & 0.0113 $\,\pm\,$ 0.00531 & 1.82 $\,\pm\,$ 0.721 \\
		\textbf{GP-Perm} & 1.27 $\,\pm\,$ 0.343 & 0.0237 $\,\pm\,$ 0.00700 & 0.0194 $\,\pm\,$ 0.00721 & 1.21 $\,\pm\,$ 0.528 \\
		DKL-DS  & 6.10 $\,\pm\,$ 3.86 & 0.00467 $\,\pm\,$ 0.00119 & 0.00568 $\,\pm\,$ 0.000942 & 5.82 $\,\pm\,$ 4.40 \\
		\bottomrule
	\end{tabular}
\end{table}

Table~\ref{tab:johansen_diagnostics} provides a complementary view of surrogate posterior and acquisition behavior during the Johansen BO loop. The posterior-standard-deviation diagnostics indicate a potential calibration issue for the representation-learning surrogates in this small-data, sequential BO regime. Both DKL and DKL-DS report very small posterior standard deviations over the fixed probe set $Z$ and at the selected candidates, while also producing large standardized residual diagnostics at the subsequently evaluated points. This combination is consistent with overconfident predictive uncertainty and/or representation drift during repeated retraining. The raw qLogEI quotient reported in the table points in the same qualitative direction, with DKL-based models showing larger and more variable relative acquisition scores, although this quantity should not be interpreted as a calibrated acquisition ratio because it is computed directly from log-acquisition values.

The table also highlights a distinction between the analytic invariant set-kernel baselines and GP-Perm. DS-GP and DE-GP generally produce lower posterior uncertainty than GP-Perm, as measured by both $\mathbb{E}_{Z}[\sigma(\bm{x})]$ and $\sigma(\bm{x}_{\text{next}})$. This behavior is plausible given their smoothing structure: DS-GP averages pairwise element similarities across sets, while DE-GP radializes an RKHS-induced set distance through an outer RBF kernel. These operations can increase similarity between distinct layouts and therefore contract posterior variance more rapidly in the few-sample regime. Consistent with this interpretation, DE-GP reports lower average posterior uncertainty than DS-GP in Table~\ref{tab:johansen_diagnostics}.

In contrast, GP-Perm compares layouts through a geometry-sensitive Sinkhorn divergence defined directly on the injector, producer, and interaction sets. In these Johansen runs, GP-Perm maintains a more moderate uncertainty profile than DS-GP and DE-GP, while also yielding a smaller standardized residual diagnostic than the other invariant GP baselines. We interpret this as evidence that GP-Perm provides a useful balance between geometric discrimination and uncertainty contraction under the tight evaluation budget. In these experiments, GP-Perm also remained numerically stable under the same standard safeguards used for the other GP models.

Overall, these diagnostics should be viewed as supporting evidence rather than as a standalone explanation of performance. Together with the AUC and final-best results, they show that GP-Perm attained the strongest sample means while maintaining moderate posterior uncertainty and less problematic standardized residual behavior than the DKL-based models in the Johansen runs. They do not isolate a causal explanation for the optimization results.

A key design choice in CCS is how to represent injector--producer coupling. Treating injectors and producers as separate exchangeable sets captures within-group structure but may miss cross-group geometry that drives operational outcomes (e.g., spacing effects and recycling risk). Our framework supports an explicit interaction representation via an interaction set constructed from injector--producer relative positions, and Appendix~C isolates the effect of including this component in both GP-Perm and DKL-DS. Conceptually, the interaction term is most valuable when the objective depends directly on injector--producer pairwise relationships rather than only on within-group properties.

Realistic CCS well placement introduces challenges that are orthogonal to permutation invariance. First, feasible well locations lie on a discrete, non-convex region induced by geology and operational constraints. Our pipeline optimizes acquisition functions in a continuous normalized space and then projects candidates to feasible grid cells while enforcing uniqueness. This projection ensures validity but can bias the induced discrete proposal distribution, which may affect exploration. Second, simulator failures can dominate experimental throughput and, in practice, determine which comparisons are possible under a fixed compute budget. The difficulty of obtaining stable trial-based results for Smeaheia in our study reinforces the value of BO pipelines that explicitly model feasibility and failure risk. Finally, while divergence-based set discrepancies are numerically stable in practice, they do not automatically imply strict positive semidefiniteness when embedded inside a radial kernel for all settings. We found standard GP stabilizers (normalization, staged jitter for Cholesky factorization, and additional safeguards reported in the diagnostics appendices) essential for robust BO operation.

\section{Conclusion}
\label{sec:conclusion}

We studied Bayesian optimization for CCS well design under group control, where the input combines unordered injector and producer location sets with auxiliary continuous controls and the objective is exactly invariant to within-group permutations. Our main contribution, \textbf{GP-Perm}, enforces this symmetry directly through an explicit permutation-invariant GP kernel built from a stable set divergence. Across a broad synthetic benchmark suite---including a CCS-like two-set objective designed to mimic injector--producer geometric coupling---and a high-fidelity CCS case study on the Johansen formation, GP-Perm provides clear mean gains on several benchmarks and achieves the highest mean AUC and mean final objective in both reported CCS settings, while remaining competitive with engineered invariant set-kernel baselines (DS/DE) and a learned-invariant DKL baseline (DKL-DS). Diagnostics on Johansen further indicate moderate posterior uncertainty and stable numerical behavior under the tight evaluation budget; the empirical comparisons are descriptive rather than claims of statistical superiority.

Our experiments suggest the following usage guidelines: (i) when exchangeability is exact and guaranteed by the problem setup (as under group control), enforcing permutation invariance can be beneficial under limited BO budgets; (ii) explicit invariant kernels are a reasonable default to evaluate when reliability under small data is critical; (iii) learned invariant embeddings can be attractive when sufficient data are available or when strict geometric assumptions are likely misaligned with the objective, but may require careful regularization and training protocols; and (iv) when wells are not truly exchangeable (e.g., well-specific constraints or controls), strict invariance may remove signal and should be relaxed.

This study focuses on fixed numbers of injectors and producers; extending the framework to variable cardinality would better reflect field development decisions. We also evaluate BO on fixed geological models; incorporating geological uncertainty (multiple realizations, calibration, or robust objectives) is needed for decision-making under subsurface uncertainty. Finally, while Deep Sets provide a strong learned-invariant baseline, more expressive set architectures (e.g., attention-based models) may better capture interaction-driven objectives, especially when combined with failure-aware acquisition and constraint handling for large-scale CCS BO campaigns.

 \section*{Credit author statement}
Conceptualization, S.F.; methodology, S.F.; software, S.F.; validation, S.F. and V.G.; resources, S.F.; writing---original draft preparation, S.F.; writing---review and editing, S.F., V.G.; visualization, S.F. All authors have read and agreed to the final version of the manuscript.

\section*{Declaration of competing interest} The authors declare that they have no known competing financial interests or personal relationships that could have appeared to influence the work reported in this paper.

\section*{Funding} The resources were granted with the support of GRNET as part of the project StorageCO$_2$.
The publication of the article in OA mode was financially supported by HEAL-Link.

\section*{Data Availability Statement} Project code can be found at https://github.com/flammmes/Permutation-Invariant-Kernels/tree/main

\bibliographystyle{elsarticle-num-names}
\bibliography{references}

\appendix
\section{Synthetic Test Functions}
\label{appendixa}
In all experiments, point coordinates are implemented in a normalized box and can be affinely mapped to $[0,1]^2$ without changing permutation-invariance; for clarity, the definitions below are given in continuous coordinates.
\subsection{Particle Configuration}
This benchmark is inspired by physics problems and tests a model's ability to reason about simple geometric interactions. It involves finding the optimal configuration of a set of $\displaystyle N$ particles in a 2-dimensional plane that minimizes a total energy function. The input is the $\displaystyle 2\cdot N$ dimensional vector representing the flattened $\displaystyle(x,y)$ coordinates of the particles.

\begin{equation}
\label{eq:prtcl}
f(\mathbb{S}) = - \left( A \cdot ||\bar{\bm{p}} - \bm{p}_{\text{target}}||^2 + B \sum_{i<j} \frac{1}{\alpha(x_i-x_j)^2 + \beta(y_i-y_j)^2 + \epsilon} \right)
\end{equation}

Here, $\mathbb{S}$ is the set of particle locations, $\displaystyle\bar{\bm{p}}$ is the centroid of the set, and $\bm{p}_{\text{target}}$ is the target location (origin). The first term is a global attractive potential pulling the set's center of mass toward the origin. The second term is an anisotropic repulsive potential that penalizes particles for being too close. Anisotropy ($\alpha \neq \beta$) is introduced to break rotational symmetry, making the problem more challenging. The objective is inherently permutation-invariant as the total energy is independent of the particle ordering. For our experiments, we used $\displaystyle N=10$ particles, resulting in a $\displaystyle 20$-dimensional search space.

\subsection{Max Area Coverage}
Inspired by sensor placement problems, this benchmark involves placing $\displaystyle N$ circles of a fixed radius $\displaystyle r$ within a $\displaystyle [-1, 1]\times[-1, 1]$ square to maximize their total union area. The objective function, which we approximate via Monte Carlo integration, presents a challenging landscape with large, nearly-flat regions and sharp gradients.

\begin{equation}
\label{eq:maxarea}
f(\mathbb{S}) = \text{Area}\left( \bigcup_{i=1}^{N} B(\bm{p}_i, r) \right) - C \sum_{i<j} \text{ReLU} \left( k \cdot (2r - ||\bm{p}_i - \bm{p}_j||) \right)
\end{equation}

In this function, $\displaystyle\mathbb{S}$ is the set of the circle centers, $B(\bm{p}_i, r)$ is the disk of radius $\displaystyle r$ centered at $\displaystyle\bm{p}_i$, and $\displaystyle \text{Area}(\cdot)$ is the area of the union of these disks. The second term is a soft penalty that discourages overlap. This objective tests a different aspect of geometric reasoning, rewarding configurations that are spread out but remain centrally located to avoid wasting coverage area outside the domain. We use $\displaystyle N=16$ circles with $\displaystyle r=0.25$, resulting in a $\displaystyle 32$-dimensional search space.

\subsection{Distribution Matching}
This benchmark is designed to be a rigorous stress test for priors that make strong geometric assumptions. The objective is purely statistical: to arrange a set of $\displaystyle N$ points such that their empirical distribution matches a fixed target distribution as closely as possible. The distance between distributions is measured by the Maximum Mean Discrepancy (MMD) \citep{gretton2012kernel}.

\begin{equation}
\label{eq:mmd}
f(\mathbb{S}) = - \max \left( 0, \frac{1}{N(N-1)}\sum_{i\neq j} k(\bm{p}_i, \bm{p}_j) + \frac{1}{M(M-1)}\sum_{i\neq j} k(\bm{y}_i, \bm{y}_j) - \frac{2}{NM}\sum_{i,j} k(\bm{p}_i, \bm{y}_j) \right)
\end{equation}

Here, $\mathbb{S} = \{\bm{p}_i\}_{i=1}^N$ is the set of candidate points and $\{\bm{y}_j\}_{j=1}^M$ is a set of $M$ fixed samples from the target distribution. The function $k(\cdot, \cdot)$ is a radial basis function (RBF) kernel. The expression inside the negation is the unbiased estimator for the squared MMD. We clamp it at zero to ensure a well-posed objective for the optimizer. The target distribution is a bimodal Gaussian mixture, requiring the model to learn both the correct locations and relative densities of the modes. For our experiments, we used $\displaystyle N=20$ points ($40$-dimensional space) and $M=5,000$ target samples.

\subsection{Max Spanning Tree}
This benchmark tests the models' ability to optimize a combinatorial property of a point set. The objective is to find a configuration of $\displaystyle N$ points in the plane that maximizes the total weight of its maximum spanning tree (MST).

\begin{equation}
\label{eq:mst}
f(\mathbb{S}) = \sum_{e \in \text{MST}(\mathbb{S})} w(e)
\end{equation}

Here, $\mathbb{S}$ is the set of points, which are treated as vertices in a complete graph. The weight $w(e)$ of an edge $e$ between two points is their Euclidean distance. $\text{MST}(\mathbb{S})$ is the set of edges that form the maximum spanning tree for the graph defined by $\mathbb{S}$. While the edge weights are geometric, the objective function is a complex, non-local property of the entire graph structure. It rewards configurations that are spread out in a structured, connected manner, providing a very different optimization landscape from the other benchmarks. For our experiments, we used $\displaystyle N=15$ points, resulting in a $\displaystyle 30$-dimensional search space.
\subsection{Facility Location (Soft Coverage)}
This benchmark is inspired by facility location and coverage objectives and tests whether the surrogate can learn a smooth \emph{soft-union} notion of coverage. The input is a set $\mathbb{S}=\{\bm{p}_i\}_{i=1}^{N}$ of $N$ two-dimensional facility locations. A fixed set of $M$ client locations $\mathbb{Y}=\{\bm{y}_j\}_{j=1}^{M}$ is sampled once and held constant throughout optimization.

We define an RBF similarity between a client $\bm{y}$ and a facility $\bm{p}$,
\begin{equation}
s(\bm{y},\bm{p})=\exp\!\left(-\frac{\|\bm{y}-\bm{p}\|_2^2}{2\sigma^2}\right),
\end{equation}
and a soft coverage probability for each client as a differentiable soft-union,
\begin{equation}
c(\bm{y};\mathbb{S}) = 1 - \prod_{i=1}^{N}\bigl(1 - s(\bm{y},\bm{p}_i)\bigr).
\end{equation}
The objective maximizes the average coverage across clients, optionally with a mild repulsion term that discourages nearly-collocated facilities:
\begin{equation}
\label{eq:facility}
f(\mathbb{S})=
\frac{1}{M}\sum_{j=1}^{M} c(\bm{y}_j;\mathbb{S})
\;-\;
\lambda \sum_{i<k} \operatorname{softplus}\!\bigl(\kappa(r_{\text{rep}}-\|\bm{p}_i-\bm{p}_k\|_2)\bigr)/\kappa .
\end{equation}
This function is permutation-invariant since it depends on $\mathbb{S}$ only through symmetric sums/products over its elements. In our experiments, $\mathbb{Y}$ is sampled either uniformly or from a simple Gaussian-mixture client distribution (fixed per benchmark instance).

\subsection{Soft $k$-Medoids}
This benchmark is a continuous, permutation-invariant relaxation of the $k$-medoids clustering objective. The input is a set $\mathbb{S}=\{\bm{p}_i\}_{i=1}^{N}$ of $N$ two-dimensional medoid locations. A fixed dataset $\mathbb{Y}=\{\bm{y}_j\}_{j=1}^{M}$ is sampled once from a prescribed distribution (e.g., Gaussian, mixture, or ring+blob) and held constant throughout optimization.

For each data point $\bm{y}$, the (hard) distance to the closest medoid is $\min_i \|\bm{y}-\bm{p}_i\|_2$. We replace the minimum with a smooth soft-min using a temperature $\tau>0$:
\begin{equation}
d_{\tau}(\bm{y};\mathbb{S})
=
-\tau \log\!\left(\sum_{i=1}^{N}\exp\!\left(-\frac{\|\bm{y}-\bm{p}_i\|_2}{\tau}\right)\right).
\end{equation}
The clustering cost is the average soft-min distance over the fixed dataset,
\begin{equation}
\label{eq:kmedoids}
\mathcal{L}(\mathbb{S}) = \frac{1}{M}\sum_{j=1}^{M} d_{\tau}(\bm{y}_j;\mathbb{S}),
\end{equation}
and the benchmark objective is defined to be maximization of the negative cost:
\begin{equation}
f(\mathbb{S}) = -\mathcal{L}(\mathbb{S}).
\end{equation}
As $\tau \to 0$, $d_{\tau}(\bm{y};\mathbb{S})$ approaches $\min_i \|\bm{y}-\bm{p}_i\|_2$, recovering a sharp $k$-medoids-like objective while remaining differentiable for $\tau>0$. The function is permutation-invariant since it depends on $\mathbb{S}$ only through symmetric operations over its elements.

\newpage
\section{Feasible Regions for CCS Benchmarks}
\label{appendixb}

\begin{figure}[h!]
    \centering
    \begin{subfigure}[b]{0.48\textwidth}
        \centering
        \includegraphics[width=0.7\linewidth]{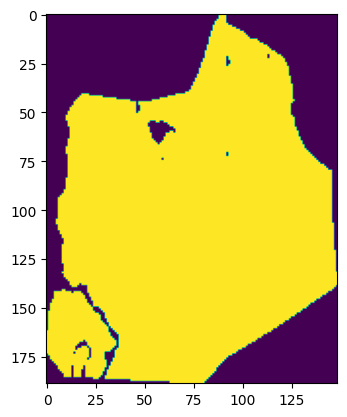}
        \caption{Initial Johansen Region}
        \label{fig:johansen_before}
    \end{subfigure}
    \hfill
    \begin{subfigure}[b]{0.48\textwidth}
        \centering
        \includegraphics[width=0.7\linewidth]{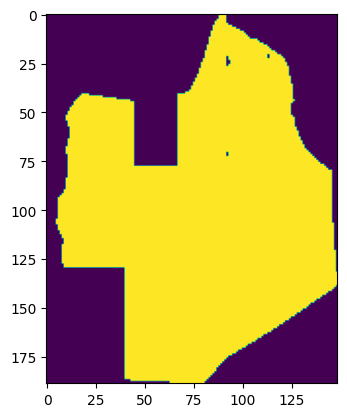}
        \caption{Final Johansen Search Space}
        \label{fig:johansen_after}
    \end{subfigure}
    \caption{
        \textbf{Feasible well placement regions for the Johansen formation.} 
        \textbf{(a)} The initial set of all valid grid cells. 
        \textbf{(b)} The final, more constrained search space after applying a mask. This mask excludes the thin western part of the formation, a step taken based on domain expertise to mitigate the risk of CO$_2$ leakage and ensure plume containment.
    }
    \label{fig:johansen_masks}
\end{figure}

\begin{figure}[h!]
    \centering
    \begin{subfigure}[b]{0.48\textwidth}
        \centering
        \includegraphics[width=0.7\linewidth]{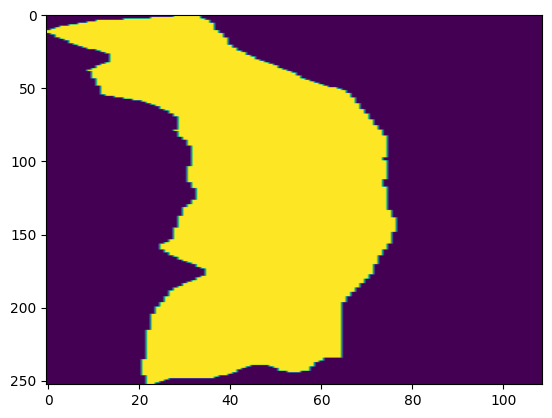}
        \caption{Initial Smeaheia Region}
        \label{fig:smeaheia_before}
    \end{subfigure}
    \hfill
    \begin{subfigure}[b]{0.48\textwidth}
        \centering
        \includegraphics[width=0.7\linewidth]{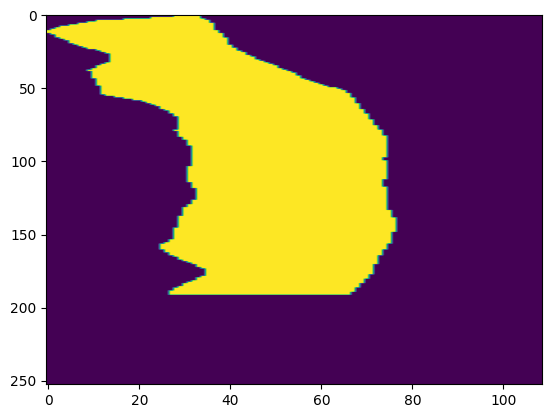}
        \caption{Final Smeaheia Search Space}
        \label{fig:smeaheia_after}
    \end{subfigure}
    \caption{
        \textbf{Feasible well placement regions for the Smeaheia formation.} 
        \textbf{(a)} The full extent of the initial valid region. 
        \textbf{(b)} The final, highly non-convex search space after applying a mask to exclude cells outside the primary fault-bounded structural trap. This refinement is crucial for ensuring that all candidate well locations are within the primary geological containment structure, preventing leakage.
    }
    \label{fig:smeaheia_masks}
\end{figure}

This appendix provides visualizations of the non-convex feasible regions for well placement in CCS formations. Figure \ref{fig:johansen_masks} represents the Johansen formation we used in this work and Figure \ref{fig:smeaheia_masks} represents the Smeaheia formation, which was not used in this work due to simulator stability issues. For each formation, we show the initial set of valid grid cells and the final, more constrained search space used by the optimizer after applying binary feasibility masks derived from domain constraints. Such constraint screening is common in CCS planning, where candidate locations are restricted to reduce leakage risk and ensure plume containment. This refinement step defines a realistic and challenging optimization problem, as discussed in Section \ref{bo_loop}.

\section{Ablation: effect of injector--producer interaction terms}
\label{appendixc}

This appendix isolates the contribution of the injector--producer interaction representation used in both GP-Perm and DKL-DS. The interaction component is designed to encode cross-group geometry (e.g., spacing, alignment, and relative positioning) that cannot be recovered from the injector and producer sets independently.

\subsection{Ablation design}
\label{appendixc:design}

We compare surrogates that are identical in all hyperparameters and training protocol, except for the inclusion of an interaction term. For GP-Perm, this corresponds to including or removing the interaction-set Sinkhorn term in the composite distance (Eq.~\ref{eq:composite_distance_ip}). For DKL-DS, this corresponds to including or removing the interaction-set Deep Sets encoder branch.

\paragraph{Models compared.}
We use the four variants defined in our experiment configuration:
\begin{itemize}
    \item \textbf{GP-Perm (full)}: interaction term enabled with weight $\texttt{ip\_weight}=1.0$.
    \item \textbf{GP-Perm (no interaction)}: interaction contribution disabled by setting $\texttt{ip\_weight}=0.0$
    (equivalently $\ell_{IP}^{-2}=0$ in Eq.~\ref{eq:composite_distance_ip}).
    \item \textbf{DKL-DS (full)}: encoder processes $\{\mathcal{I},\mathcal{P},\mathcal{R}\}$ with $\texttt{rel\_weight}=1.0$.
    \item \textbf{DKL-DS (no interaction)}: interaction branch is effectively removed by setting $\texttt{rel\_weight}=0.0$,
    so the representation depends on $\{\mathcal{I},\mathcal{P}\}$ only.
\end{itemize}

\paragraph{Controlled hyperparameters.}
All remaining settings (kernel type, optimizer, learning rate, jitter schedule, acquisition function, and BO budget) are held fixed within each model family and match the protocol used for this synthetic two-set experiment. In particular, the injector/producer counts are fixed to $n_{\mathrm{inj}}=4$ and $n_{\mathrm{prod}}=6$, and each configuration is evaluated over $n_{\mathrm{trials}}=10$ independent trials.\\

\subsection{Synthetic two-set benchmark (explicit objective)}
\label{appendixc:benchmark}

To emphasize cross-group coupling, we use a synthetic objective that rewards \emph{good matching} between producers and injectors
while preventing within-group collapse via repulsion. The decision variable is a flattened vector
\[
\bm{x}\in[-1,1]^{2(n_{\mathrm{inj}}+n_{\mathrm{prod}})} ,
\]
which is reshaped into $(n_{\mathrm{inj}}+n_{\mathrm{prod}})$ planar points. We write
$\mathcal{I}=\{\bm{i}_a\}_{a=1}^{n_{\mathrm{inj}}}$ and $\mathcal{P}=\{\bm{p}_b\}_{b=1}^{n_{\mathrm{prod}}}$, where $\bm{i}_a,\bm{p}_b\in\mathbb{R}^2$.
As in Eq.~\ref{eq:interaction_set}, the interaction set is $\mathcal{R}=\{\bm{p}_b-\bm{i}_a\}_{a,b}$.

Define squared pairwise distances
\[
d_{ba}^2 = \|\bm{p}_b-\bm{i}_a\|_2^2 .
\]
For each producer, we compute a smooth approximation of the minimum distance to an injector using a soft-min with temperature $\tau>0$:
\begin{equation}
\label{eq:twoset_softmin}
s_b(\mathcal{I},\mathcal{P})
=
-\tau \log \sum_{a=1}^{n_{\mathrm{inj}}}\exp\!\left(-\frac{d_{ba}^2}{\tau}\right).
\end{equation}
The interaction cost averages this quantity over producers:
\begin{equation}
\label{eq:twoset_interaction_cost}
C_{\mathcal{IP}}(\mathcal{I},\mathcal{P})
=
\frac{1}{n_{\mathrm{prod}}}\sum_{b=1}^{n_{\mathrm{prod}}} s_b(\mathcal{I},\mathcal{P}).
\end{equation}
As $\tau\rightarrow 0$, $s_b$ approaches $\min_a d_{ba}^2$, i.e., each producer is encouraged to be close to some injector.

To avoid degenerate solutions where all points collapse, we add inverse-distance repulsion within each set:
\begin{equation}
\label{eq:twoset_repulsion}
C_{\mathrm{rep}}(\mathcal{I},\mathcal{P})
=
\frac{\lambda_{\mathrm{inj}}}{2}\sum_{\substack{a,a'=1\\a\neq a'}}^{n_{\mathrm{inj}}}
\frac{1}{\|\bm{i}_a-\bm{i}_{a'}\|_2^2+\varepsilon}
\;+\;
\frac{\lambda_{\mathrm{prod}}}{2}\sum_{\substack{b,b'=1\\b\neq b'}}^{n_{\mathrm{prod}}}
\frac{1}{\|\bm{p}_b-\bm{p}_{b'}\|_2^2+\varepsilon},
\end{equation}
with a small $\varepsilon>0$ for numerical stability.

The benchmark defines a total cost
\begin{equation}
\label{eq:twoset_total_cost}
C(\mathcal{I},\mathcal{P}) = C_{\mathcal{IP}}(\mathcal{I},\mathcal{P}) + C_{\mathrm{rep}}(\mathcal{I},\mathcal{P}),
\end{equation}
and the scalar objective for BO is \emph{maximization of the negative cost}:
\begin{equation}
\label{eq:twoset_objective}
f(\mathcal{I},\mathcal{P}) = -\,C(\mathcal{I},\mathcal{P}).
\end{equation}
We use the defaults $\tau=0.05$, $\lambda_{\mathrm{inj}}=\lambda_{\mathrm{prod}}=0.05$, and $\varepsilon=10^{-4}$,
with $n_{\mathrm{inj}}=4$ and $n_{\mathrm{prod}}=6$.

\subsection{BO protocol and metrics}
\label{appendixc:protocol}

We run BO using the same candidate parameterization and best-so-far reporting as in the main experiments, specialized to this
two-set synthetic objective.

Each trial starts from $n_{\mathrm{init}}=6$ randomly drawn points in the bounded domain, followed by $T=25$ BO iterations with batch size $q=4$,
for a total of $n_{\mathrm{eval}}=n_{\mathrm{init}} + Tq = 106$ function evaluations per trial.

We use qLogExpectedImprovement (qLogEI) with a fixed Monte Carlo sampler and optimize the acquisition function in the normalized domain
$[0,1]^d$ using multi-start optimization (10 restarts and 256 raw samples). The evaluation points are then unnormalized back to $[-1,1]^d$
and scored by Eq.~\ref{eq:twoset_objective}.

We report (i) mean best-so-far trajectories across trials, (ii) AUC of best-so-far curves, and (iii) final best-so-far values.
All metrics are aggregated over the $n_{\mathrm{trials}}=10$ trials for this benchmark instance.

\subsection{Results}
\label{appendixc:results}

\begin{figure}[t!]
\centering
\includegraphics[width=0.95\linewidth]{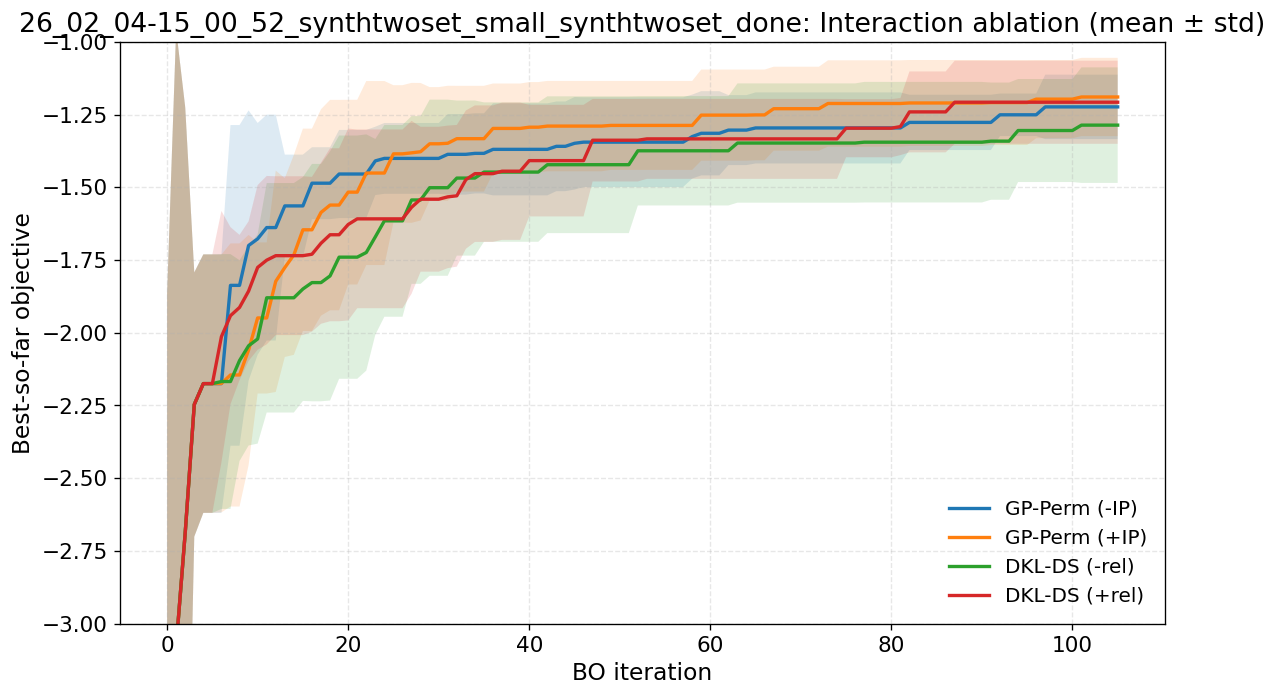}
\caption{Interaction-term ablation on the synthetic two-set benchmark (Eq.~\ref{eq:twoset_objective}).
Mean best-so-far trajectories across $n_{\mathrm{trials}}=10$ trials for GP-Perm and DKL-DS, with and without interaction modeling (shaded: $\pm 1$ std).}
\label{fig:appendixC_best_so_far}
\end{figure}

Figure~\ref{fig:appendixC_best_so_far} compares the mean best-so-far trajectories for GP-Perm and DKL-DS with and without explicit interaction modeling.
Table~\ref{tab:appendixC_summary} reports AUC and final best-so-far values (mean $\pm$ std across trials). Since the objective is $f=-C$ (Eq.~\ref{eq:twoset_objective}),
values closer to zero indicate better solutions.

Overall, including the interaction term improves performance in both model families under the fixed BO budget.
For GP-Perm, enabling the interaction component yields a small improvement in both AUC and final best-so-far.
For DKL-DS, the interaction branch provides a larger mean uplift, consistent with the purpose of the relational representation, although variance across trials remains non-negligible.

\begin{table}[t!]
\centering
\caption{Interaction-term ablation summary on the synthetic two-set benchmark (Eq.~\ref{eq:twoset_objective}).
Report mean$\pm$std across $n_{\mathrm{trials}}=10$ trials for AUC of best-so-far curves and final best-so-far (higher is better). Best per metric is bolded.}
\label{tab:appendixC_summary}
\begin{tabular}{@{}lcc@{}}
\toprule
\textbf{Model} & \textbf{AUC (mean$\pm$std)} & \textbf{Final best-so-far (mean$\pm$std)} \\
\midrule
GP-Perm (full; +IP) & \textbf{-149.052 $\pm$ 17.817} & \textbf{-1.190 $\pm$ 0.134} \\
GP-Perm (no interaction; -IP) & -150.787 $\pm$ 9.013 & -1.224 $\pm$ 0.111 \\
DKL-DS (full; +rel) & -155.230 $\pm$ 12.525 & -1.208 $\pm$ 0.143 \\
DKL-DS (no interaction; -rel) & -161.654 $\pm$ 23.340 & -1.286 $\pm$ 0.199 \\
\bottomrule
\end{tabular}
\end{table}

This ablation directly tests whether the surrogate needs an explicit cross-group representation to optimize a function whose primary
signal is induced by injector--producer coupling (Eqs.~\ref{eq:twoset_softmin}--\ref{eq:twoset_objective}). Improvements in AUC indicate higher sample-efficiency, while improvements in final best-so-far indicate better asymptotic solution quality under the fixed budget.

\section{PSD conditions, finite-design guarantees, and kernel geometry diagnostics for GP-Perm}
\label{appendixd}

Positive semidefiniteness has two distinct meanings in the present setting. A universal Mercer guarantee requires a positive-semidefinite (PSD) Gram matrix for every finite collection of admissible inputs and every permitted hyperparameter value. Operational GP inference, in contrast, only factorizes the finite matrices generated along a particular BO trajectory. Symmetry, non-negativity, and vanishing self-divergence of $S_{\varepsilon}$ do not alone imply that $S_{\varepsilon}$ is of negative type, which is the standard route from a squared dissimilarity to a radial PSD kernel. We therefore do not claim an unconditional Mercer theorem for finite-$\varepsilon$ GP-Perm. Instead, this appendix gives (i) a transparent sufficient condition for such a theorem, (ii) deterministic finite-design conditions that do not require negative type, (iii) a perturbation guarantee around every matrix with a positive spectral margin, (iv) a simulator-free study of invariance, continuity, regularization, and interaction geometry, and (v) spectral diagnostics on the finite matrices used in the original experiments. These statements apply to both the single-set construction and the full two-set-plus-auxiliary construction.

\subsection{Notation and the matrix used by GP inference}
\label{appendixd:notation}

For a finite design $\mathcal{X}_N=\{\bm{x}_1,\ldots,\bm{x}_N\}$, let
\begin{equation}
\Psi_{ij}=D^2(\bm{x}_i,\bm{x}_j),
\qquad
K_{ij}=\sigma^2\rho_{\nu}\!\left(D(\bm{x}_i,\bm{x}_j)\right),
\label{eq:appendixd_gram}
\end{equation}
where $\rho_{\nu}$ is the unit-variance Mat\'ern correlation function,
\begin{equation}
\rho_{\nu}(r)
=
\frac{2^{1-\nu}}{\Gamma(\nu)}
\bigl(\sqrt{2\nu}\,r\bigr)^{\nu}
K_{\nu}^{\mathrm{Bessel}}\!\bigl(\sqrt{2\nu}\,r\bigr),
\qquad \rho_{\nu}(0)=1.
\label{eq:appendixd_matern}
\end{equation}
Here $K_{\nu}^{\mathrm{Bessel}}$ denotes the modified Bessel function of the second kind. The function $\rho_{\nu}$ is positive, continuous, decreasing on $[0,\infty)$, and converges to zero as $r\to\infty$. For the Mat\'ern-$5/2$ kernel used in GP-Perm,
\begin{equation}
\rho_{5/2}(r)
=
\left(1+\sqrt{5}\,r+\frac{5}{3}r^2\right)e^{-\sqrt{5}r}.
\label{eq:appendixd_matern52}
\end{equation}
The matrix factorized during GP training can be written as
\begin{equation}
C=K+\eta I_N,
\qquad
\eta=\sigma_n^2+\tau\geq 0,
\label{eq:appendixd_factorized}
\end{equation}
where $\sigma_n^2$ is the Gaussian likelihood variance and $\tau$ is the numerical jitter. The diagnostics below deliberately examine the more stringent raw kernel matrix $K$; hence $\eta=0$ for the reported eigenvalues unless explicitly stated otherwise.

\subsection{A conditional global PSD criterion}
\label{appendixd:global}

A symmetric function $q$ with $q(z,z)=0$ is \emph{conditionally negative semidefinite} (CND), or of negative type, if for every finite collection $z_1,\ldots,z_N$ and every $\bm{c}\in\mathbb{R}^N$ satisfying $\bm{1}^{\top}\bm{c}=0$,
\begin{equation}
\sum_{i=1}^{N}\sum_{j=1}^{N}c_i c_j q(z_i,z_j)\leq 0.
\label{eq:appendixd_cnd}
\end{equation}

\paragraph{Proposition D.1 (conditional global guarantee).}
Suppose that $S_{\varepsilon}$ is CND on each set domain on which it is used. Then the squared composite dissimilarity $D^2$ in Eq.~\ref{eq:composite_distance_ip} is CND, and $k_{\mathrm{GP-Perm}}=\sigma^2\rho_{\nu}\circ D$ is PSD for every $\nu>0$, every positive collection of lengthscales, and every finite design. The same statement holds for the single-set special case.

\emph{Proof.}
The ARD term is CND because, for coefficients summing to zero,
\begin{equation}
\sum_{i,j}c_i c_j
\left\|\bm{L}^{-1}(\bm{v}_i-\bm{v}_j)\right\|_2^2
=-2\left\|\sum_i c_i\bm{L}^{-1}\bm{v}_i\right\|_2^2\leq 0.
\end{equation}
Positive rescaling and addition preserve the CND property. Moreover, the interaction term is a pullback of $S_{\varepsilon}$ through the deterministic map $(\mathcal I,\mathcal P)\mapsto\mathcal R$ and therefore is CND whenever $S_{\varepsilon}$ is CND on the interaction-set domain. Hence $\Psi=D^2$ is CND. Schoenberg's theorem then gives that $[e^{-t\Psi_{ij}}]_{ij}$ is PSD for every $t>0$ \citep{schoenberg1938metric}. Finally, the Mat\'ern correlation admits the positive Gaussian-mixture representation
\begin{equation}
\rho_{\nu}(\sqrt{\Psi})
=
\int_{0}^{\infty}e^{-t\Psi}
\frac{(\nu/2)^{\nu}}{\Gamma(\nu)}
t^{-\nu-1}\exp\!\left(-\frac{\nu}{2t}\right)\,\mathrm{d}t.
\label{eq:appendixd_matern_mixture}
\end{equation}
where the square root and exponential are applied entrywise. It is therefore a non-negative mixture of PSD matrices, proving the claim.

Proposition D.1 isolates the exact missing premise in a universal argument: a non-negative Sinkhorn divergence is not automatically CND at finite regularization. The proposition is thus a sufficient criterion, not an assertion that this premise holds over the unrestricted GP-Perm input space.

There is also an exact finite-design version of this criterion. Let
\begin{equation}
H=I_N-\frac{1}{N}\bm{1}\bm{1}^{\top}.
\end{equation}
The restriction of $D^2$ to $\mathcal X_N$ is CND if and only if
\begin{equation}
H\Psi H\preceq 0.
\label{eq:appendixd_finite_cnd}
\end{equation}
Whenever Eq.~\ref{eq:appendixd_finite_cnd} holds, Eq.~\ref{eq:appendixd_matern_mixture} certifies the Mat\'ern Gram matrix on that design for every $\nu>0$ and every common positive rescaling of $D^2$. This condition is stronger than needed for one particular fitted Mat\'ern matrix, so the next result supplies a direct certificate for that matrix.

\subsection{An unconditional finite-design row-sum certificate}
\label{appendixd:rowsum}

\paragraph{Proposition D.2 (diagonal-dominance guarantee).}
For any symmetric, non-negative dissimilarity matrix $[D_{ij}]$ with $D_{ii}=0$, define
\begin{equation}
g_i=\sum_{j\neq i}\rho_{\nu}(D_{ij}),
\qquad
g_{\star}=\max_{1\leq i\leq N}g_i.
\label{eq:appendixd_rowsum}
\end{equation}
Then the factorized GP matrix in Eq.~\ref{eq:appendixd_factorized} satisfies
\begin{equation}
\lambda_{\min}(C)
\geq
\sigma^2(1-g_{\star})+\eta.
\label{eq:appendixd_gershgorin}
\end{equation}
Consequently,
\begin{equation}
g_{\star}<1+\frac{\eta}{\sigma^2}
\label{eq:appendixd_rowsum_condition}
\end{equation}
is sufficient for $C$ to be strictly positive definite. Replacing $<$ by $\leq$ gives a sufficient PSD condition.

\emph{Proof.}
The diagonal of $C$ is $\sigma^2+\eta$, while the absolute off-diagonal row sum is $\sigma^2 g_i$. Equation~\ref{eq:appendixd_gershgorin} follows directly from the Gershgorin circle theorem and the fact that $C$ is symmetric.

A more conservative condition depends only on the smallest pairwise composite separation,
\begin{equation}
D_{\min}=\min_{i\neq j}D_{ij}.
\end{equation}
Since $\rho_{\nu}$ is decreasing, $g_{\star}\leq(N-1)\rho_{\nu}(D_{\min})$. Hence
\begin{equation}
(N-1)\rho_{\nu}(D_{\min})
<1+\frac{\eta}{\sigma^2}
\label{eq:appendixd_dmin_condition}
\end{equation}
is sufficient for strict positive definiteness. For the Mat\'ern-$5/2$ kernel, this becomes the fully explicit condition
\begin{equation}
(N-1)
\left(1+\sqrt{5}D_{\min}+\frac{5}{3}D_{\min}^2\right)
e^{-\sqrt{5}D_{\min}}
<1+\frac{\eta}{\sigma^2}.
\label{eq:appendixd_matern52_condition}
\end{equation}
No CND, Hilbert-embedding, or triangle-inequality assumption is used in Proposition D.2.

For a single set without auxiliary variables, $D_{ij}^2=S_{\varepsilon}(S_i,S_j)/\ell_S^2$, and Eqs.~\ref{eq:appendixd_rowsum_condition}--\ref{eq:appendixd_matern52_condition} apply directly. For the full construction, $D_{ij}^2$ is the sum of the auxiliary, injector, producer, and interaction contributions in Eq.~\ref{eq:composite_distance_ip}. Holding the existing components and their hyperparameters fixed, adding any non-negative component increases each $D_{ij}$, weakly decreases every off-diagonal correlation, and therefore cannot weaken the row-sum certificate. In particular, adding the auxiliary or interaction block may make Eq.~\ref{eq:appendixd_rowsum_condition} easier to satisfy. This monotonicity concerns the certificate under fixed shared hyperparameters; it does not assert that the minimum eigenvalue must increase when two variants are fitted separately and learn different lengthscales.

\subsection{A non-empty positive-definite lengthscale region}
\label{appendixd:basin}

\paragraph{Proposition D.3 (existence of an SPD basin).}
Fix a finite collection of designs that are distinct on the quotient space induced by within-set permutations, and suppose that every pair is separated by at least one unscaled component of $D^2$. Fix positive reference lengthscales $\{\ell_b^{(0)}\}$ for the active blocks and set $\ell_b(a)=a\ell_b^{(0)}$ for $a>0$. Then there exists $a_0>0$ such that the raw kernel matrix $K(a)$ is strictly positive definite for every $0<a<a_0$.

\emph{Proof.}
Under the common rescaling, $D_{ij}(a)=D_{ij}(1)/a$. Finiteness and quotient-distinctness imply $d_{\star}=\min_{i\neq j}D_{ij}(1)>0$. Therefore
\begin{equation}
\rho_{\nu}(D_{ij}(a))\leq \rho_{\nu}(d_{\star}/a)\longrightarrow 0
\qquad\text{as }a\downarrow 0.
\end{equation}
It follows that $K(a)\to\sigma^2I_N$. Equivalently, for sufficiently small $a$, $(N-1)\rho_{\nu}(d_{\star}/a)<1$, and Proposition D.2 applies with $\eta=0$. Because the inequality is strict and the eigenvalues vary continuously with $a$, the valid set is open.

Thus, even without a global CND property, every finite quotient-distinct GP-Perm design has a non-empty, open lengthscale region in which its Gram matrix is strictly positive definite. If exact duplicate designs are present, the raw matrix necessarily has repeated rows and cannot be strictly positive definite. After grouping duplicates, however, the small-$a$ limit consists of PSD all-ones blocks; consequently, any fixed $\eta>0$ makes the factorized matrix strictly positive definite for all sufficiently small $a$.

\subsection{Stability under numerical and hyperparameter perturbations}
\label{appendixd:stability}

\paragraph{Proposition D.4 (spectral-margin guarantee).}
Let an audited symmetric kernel matrix $K$ have $\lambda_{\min}(K)=\gamma$, and let $\widehat K=K+E$ for a symmetric perturbation $E$. Then
\begin{equation}
\lambda_{\min}(\widehat K+\eta I_N)
\geq \gamma+\eta-\|E\|_2.
\label{eq:appendixd_weyl}
\end{equation}
Hence every matrix satisfying $\|E\|_2<\gamma+\eta$ remains strictly positive definite after the diagonal term is included.

\emph{Proof.}
Equation~\ref{eq:appendixd_weyl} is Weyl's eigenvalue perturbation inequality. If the entries obey $|E_{ij}|\leq e$, the conservative bound $\|E\|_2\leq Ne$ gives the readily checked sufficient condition $Ne<\gamma+\eta$.

For the Mat\'ern-$5/2$ profile, this can also be expressed in terms of distance perturbations. Differentiation of Eq.~\ref{eq:appendixd_matern52} gives
\begin{equation}
\left|\rho'_{5/2}(r)\right|
=\frac{5}{3}r(1+\sqrt{5}r)e^{-\sqrt{5}r}
\leq L_{5/2}<0.63,
\label{eq:appendixd_lipschitz}
\end{equation}
where the maximum occurs at $\sqrt{5}r=(1+\sqrt{5})/2$. If the output scale is fixed and $|\widehat D_{ij}-D_{ij}|\leq\delta_D$ for all pairs, the mean-value theorem gives $|E_{ij}|\leq0.63\sigma^2\delta_D$. A sufficient local robustness condition is therefore
\begin{equation}
0.63N\sigma^2\delta_D<\gamma+\eta.
\label{eq:appendixd_distance_robustness}
\end{equation}
Proposition D.4 formalizes the fact that a positive observed eigenvalue is not a knife-edge event: it certifies an open neighborhood of inputs, distances, and numerical evaluations whose radius is controlled by the observed spectral margin.

\subsection{Simulator-free kernel sanity and geometry study}
\label{appendixd:geometry}

We next examine GP-Perm independently of surrogate fitting or optimization. This study uses no simulator evaluations and no previously collected BO outcomes: every input is a newly generated injector--producer configuration in the normalized domain $[0,1]^2$. The purpose is to audit the geometry supplied by the kernel itself.

\paragraph{Protocol and scale matching.}
We use $n_{\mathrm{inj}}=4$, $n_{\mathrm{prod}}=6$, the Mat\'ern-$5/2$ profile, reference regularization $\varepsilon=0.05$, and reference common block lengthscale $\ell=1.5$. An independent NumPy implementation computes the uniform-marginal entropy-regularized coupling and, consistently with the GP-Perm implementation, reports its sharp transport cost $\langle P,C\rangle$ before applying the debiasing in Eq.~\ref{eq:sinkhorn_divergence} and the non-negativity clamp. The maximum marginal residual over the entire study is $8.00\times10^{-6}$.

The three set blocks have different cardinalities and natural scales. We therefore form fixed diagnostic scales
\begin{equation}
s_c=\operatorname{median}_{q<r}\,S_{\varepsilon_0}\!\left(Z_q^{(c)},Z_r^{(c)}\right),
\qquad
\widetilde S_c=S_c/s_c,
\qquad c\in\{I,P,R\},
\label{eq:appendixd_diagnostic_scaling}
\end{equation}
from 512 random calibration pairs at $\varepsilon_0=0.05$ and reuse them for every subsequent setting. Here $Z_q^{(c)}$ denotes block $c$ of design $q$. This nondimensionalization is exactly absorbable into the block lengthscales ($\ell_c\leftarrow\ell\sqrt{s_c}$), so it does not introduce a different kernel family. To prevent arbitrary baseline lengthscales from deciding the perturbation comparison, the DS-GP element lengthscale and the DE-GP outer lengthscale are calibrated once so that their median normalized correlation over random layouts equals the reference GP-Perm median, $0.438$. All reported similarities are normalized kernel correlations $k(x,x')/\sqrt{k(x,x)k(x',x')}$.

\begin{figure}[t!]
\centering
\includegraphics[width=\linewidth]{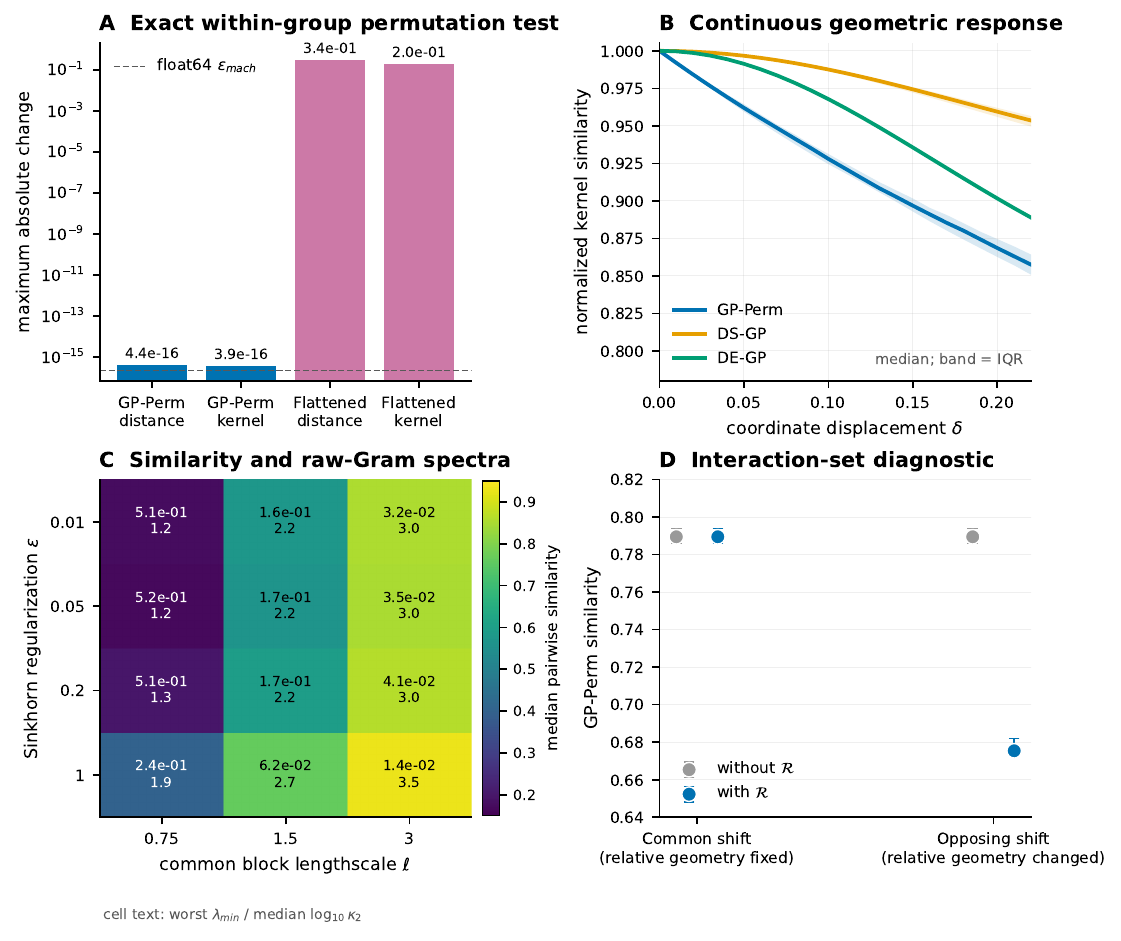}
\caption{Simulator-free GP-Perm kernel sanity and geometry study. \textbf{A:} maximum change over 160 random design pairs after independently permuting injector and producer order in one argument. \textbf{B:} similarity when one injector and one producer are displaced along independent random directions (median and interquartile range over 64 base layouts); DS-GP and DE-GP are median-similarity matched at the reference setting. \textbf{C:} sensitivity of two-set GP-Perm without $\mathcal R$ over 12 random $48\times48$ raw Gram matrices per cell. Color is median off-diagonal similarity; cell text gives the worst $\lambda_{\min}$ and median $\log_{10}\kappa_2$. \textbf{D:} controlled comparison of a common translation and an opposing injector--producer translation (median and interquartile range over 72 layouts). This is a kernel-geometry diagnostic, not a BO-performance comparison.}
\label{fig:appendixD_geometry}
\end{figure}

\paragraph{Exact invariance and continuous perturbations.}
For 160 random design pairs, independently permuting injector and producer rows changes the full GP-Perm distance by at most $4.44\times10^{-16}$ and its kernel value by at most $3.89\times10^{-16}$ (Figure~\ref{fig:appendixD_geometry}A). These are float64 rounding-level discrepancies. Under the identical permutations, a Mat\'ern-$5/2$ kernel on the flattened coordinate vector changes its normalized distance by as much as $0.336$ and its kernel value by as much as $0.204$. Thus the numerical check distinguishes invariance of the implementation from invariance only of the underlying objective.

For Figure~\ref{fig:appendixD_geometry}B, one injector and one producer in each base design are displaced by $\delta\in[0,0.22]$ along independent random directions. All three set-aware kernels change continuously at the resolution tested. GP-Perm decreases smoothly and, under the matched random-layout calibration above, retains more local geometric discrimination: at $\delta=0.22$, its median similarity is $0.857$, compared with $0.889$ for DE-GP and $0.954$ for DS-GP. The result is a statement about the induced geometry under this controlled protocol.

\paragraph{Regularization, lengthscale, and finite Gram matrices.}
To isolate the $\varepsilon$--$\ell$ effect from the explicit interaction representation, Figure~\ref{fig:appendixD_geometry}C uses the injector and producer blocks; Figure~\ref{fig:appendixD_geometry}D then audits $\mathcal R$ separately. We generate 12 independent sets of $N=48$ designs and sweep $\varepsilon\in\{0.01,0.05,0.2,1\}$ and $\ell\in\{0.75,1.5,3\}$ while holding the scales in Eq.~\ref{eq:appendixd_diagnostic_scaling} fixed. This exposes rather than renormalizes away the transition from an OT-like small-$\varepsilon$ comparison to a more pooled, MMD-like large-$\varepsilon$ comparison.

All 144 raw Gram matrices have $\lambda_{\min}>0$; the worst value over the complete grid is $1.3578\times10^{-2}$. The sweep also identifies the soft operating tradeoff. At large $\varepsilon$ and $\ell$, distinct layouts become highly correlated: for $(\varepsilon,\ell)=(1,3)$ the median pairwise similarity is $0.928$, the worst $\lambda_{\min}$ is $1.36\times10^{-2}$, and the median condition number is $3.01\times10^3$. In contrast, at $(0.05,0.75)$ these values are $0.171$, $0.521$, and $16.3$, respectively. Positive definiteness was retained throughout this grid, but similarity saturation predictably reduced the spectral margin and worsened conditioning. Table~\ref{tab:appendixD_geometry_sensitivity} gives the regularization slice at the reference lengthscale.

\begin{table}[t!]
\centering
\scriptsize
\setlength{\tabcolsep}{3.5pt}
\caption{Simulator-free regularization sensitivity at the reference block lengthscale $\ell=1.5$ ($N=48$, 12 random Gram matrices per setting). Similarity statistics are the medians across replicates; $\lambda_{\min}$ is the worst raw-matrix value; $\kappa_2$ is the median raw 2-norm condition number. A violation means $\lambda_{\min}<-10^{-10}$.}
\label{tab:appendixD_geometry_sensitivity}
\begin{tabular}{ccccc}
\toprule
$\varepsilon$ & Similarity: median [P10, P90] & Worst $\lambda_{\min}$ & Median $\kappa_2$ & Violations \\
\midrule
0.01 & 0.573 [0.504, 0.638] & $1.63\times 10^{-1}$ & $1.61\times 10^{2}$ & 0/12 \\
0.05 & 0.566 [0.498, 0.630] & $1.68\times 10^{-1}$ & $1.53\times 10^{2}$ & 0/12 \\
0.2 & 0.595 [0.530, 0.660] & $1.70\times 10^{-1}$ & $1.63\times 10^{2}$ & 0/12 \\
1 & 0.760 [0.677, 0.822] & $6.20\times 10^{-2}$ & $5.46\times 10^{2}$ & 0/12 \\
\bottomrule
\end{tabular}
\end{table}

For every matrix in this finite suite, the computed eigendecomposition is a direct PSD certificate. Moreover, Proposition D.4 turns the smallest observed margin into a suite-wide local guarantee: every audited raw matrix remains positive definite under any symmetric perturbation with spectral norm below $1.3578\times10^{-2}$. Likelihood noise or jitter enlarges its individual margin additively. This is deliberately a finite-design robustness statement; the random sweep cannot establish a universal Mercer theorem.

\paragraph{Interaction-set diagnostic.}
Finally, let $u$ be a random unit vector and $a=0.12$. Relative to a base layout, the common-shift design uses $(I+au,P+au)$, while the opposing-shift design uses $(I+au,P-au)$. The injector and producer marginal comparisons have identical geometry in the two cases, so GP-Perm without $\mathcal R$ gives the same median similarity, $0.790$. Under the common shift, $\mathcal R$ is unchanged and adding it also leaves the median at $0.790$. Under the opposing shift, however, $\mathcal R$ is translated by $-2au$ and the full-kernel median falls to $0.675$ (Figure~\ref{fig:appendixD_geometry}D). The controlled construction shows exactly what the interaction block contributes: it distinguishes a change in injector--producer alignment that is invisible to the two matched marginal comparisons.

\subsection{Offline PSD stress test on random designs}
\label{appendixd:offline}

The following retained results concern the synthetic two-set experiment in Appendix~\ref{appendixc}, with $d=2(n_{\mathrm{inj}}+n_{\mathrm{prod}})=20$, and use learned GP-Perm hyperparameters from a representative trained BO run. Unlike the preceding simulator-free geometry study, these settings are tied to the original experiments. We draw random point sets $\{\bm{x}_i\}_{i=1}^N$ uniformly in $[0,1]^d$ and compute the raw kernel matrix $K$ in Eq.~\ref{eq:appendixd_gram}. We report $\lambda_{\min}(K)$ and the fraction of draws where $\lambda_{\min}(K)<-\delta$ for $\delta\in\{10^{-10},10^{-8},10^{-6}\}$. We test $N\in\{64,128\}$ with $20$ independent draws per $N$ and per variant.

\begin{figure}[t!]
\centering
\includegraphics[width=0.95\linewidth]{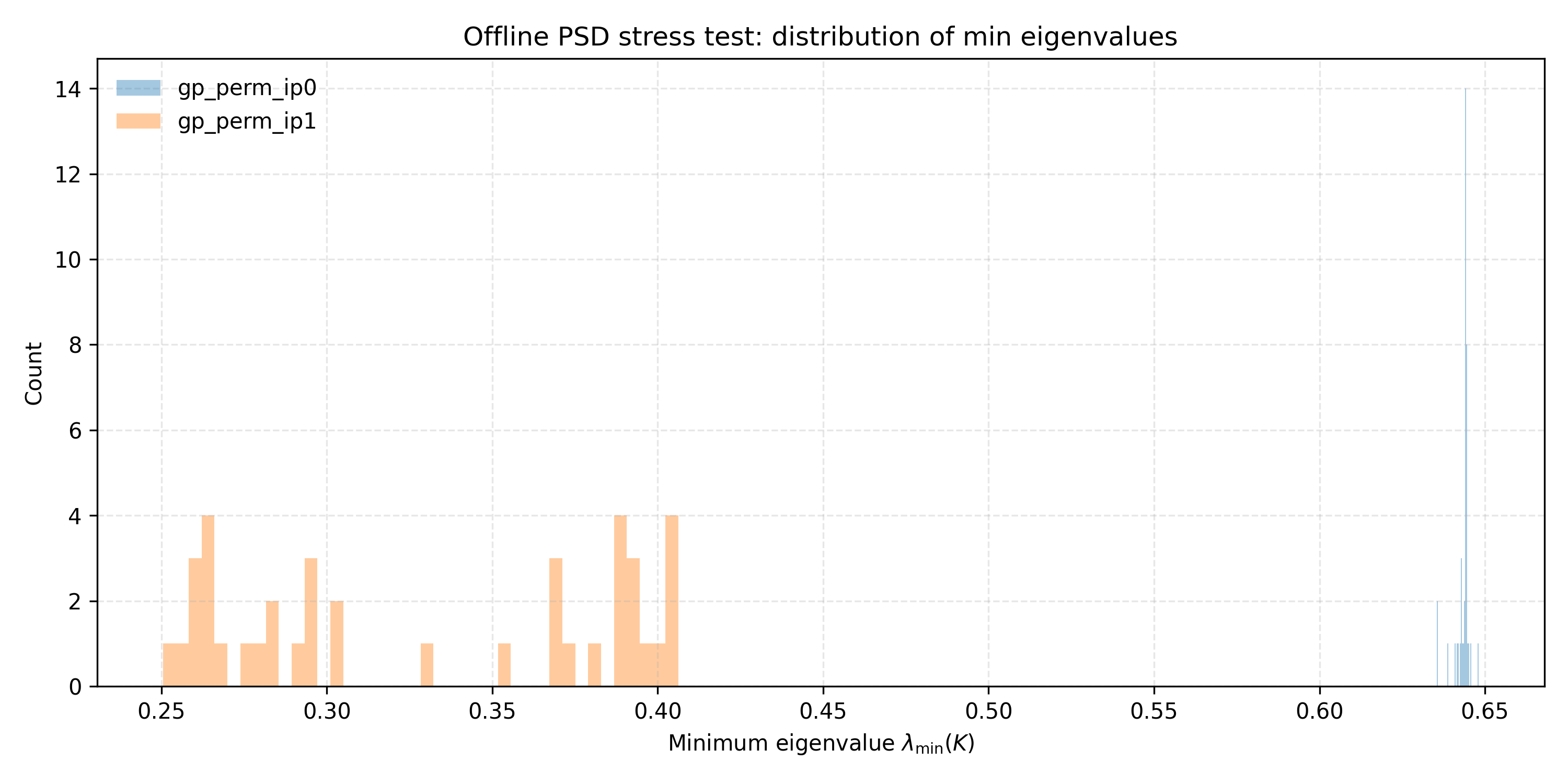}
\caption{Offline PSD stress test: histogram of $\lambda_{\min}(K)$ across random designs in $[0,1]^d$.
Both GP-Perm variants produce positive minimum eigenvalues in all tested draws.}
\label{fig:appendixD_offline_hist}
\end{figure}

\begin{table}[t!]
\centering
\small
\caption{Offline PSD stress test summary (random designs in $[0,1]^d$). We report median / mean $\lambda_{\min}(K)$ and the fraction of draws where $\lambda_{\min}(K)<-\delta$.}
\label{tab:appendixD_offline_psd}
\begin{tabular}{lcccccc}
\toprule
\textbf{Variant} & $N$ & $\mathrm{median}\,\lambda_{\min}$ & $\mathrm{mean}\,\lambda_{\min}$ &
$\delta=10^{-10}$ & $\delta=10^{-8}$ & $\delta=10^{-6}$\\
\midrule
GP-Perm (-IP) & 64  & 0.644 & 0.644 & 0.000 & 0.000 & 0.000 \\
GP-Perm (-IP) & 128 & 0.644 & 0.643 & 0.000 & 0.000 & 0.000 \\
GP-Perm (+IP) & 64  & 0.390 & 0.384 & 0.000 & 0.000 & 0.000 \\
GP-Perm (+IP) & 128 & 0.272 & 0.276 & 0.000 & 0.000 & 0.000 \\
\bottomrule
\end{tabular}
\end{table}

\subsection{Training-set PSD diagnostic along BO trajectories}
\label{appendixd:training}

\begin{figure}[t!]
\centering
\includegraphics[width=0.95\linewidth]{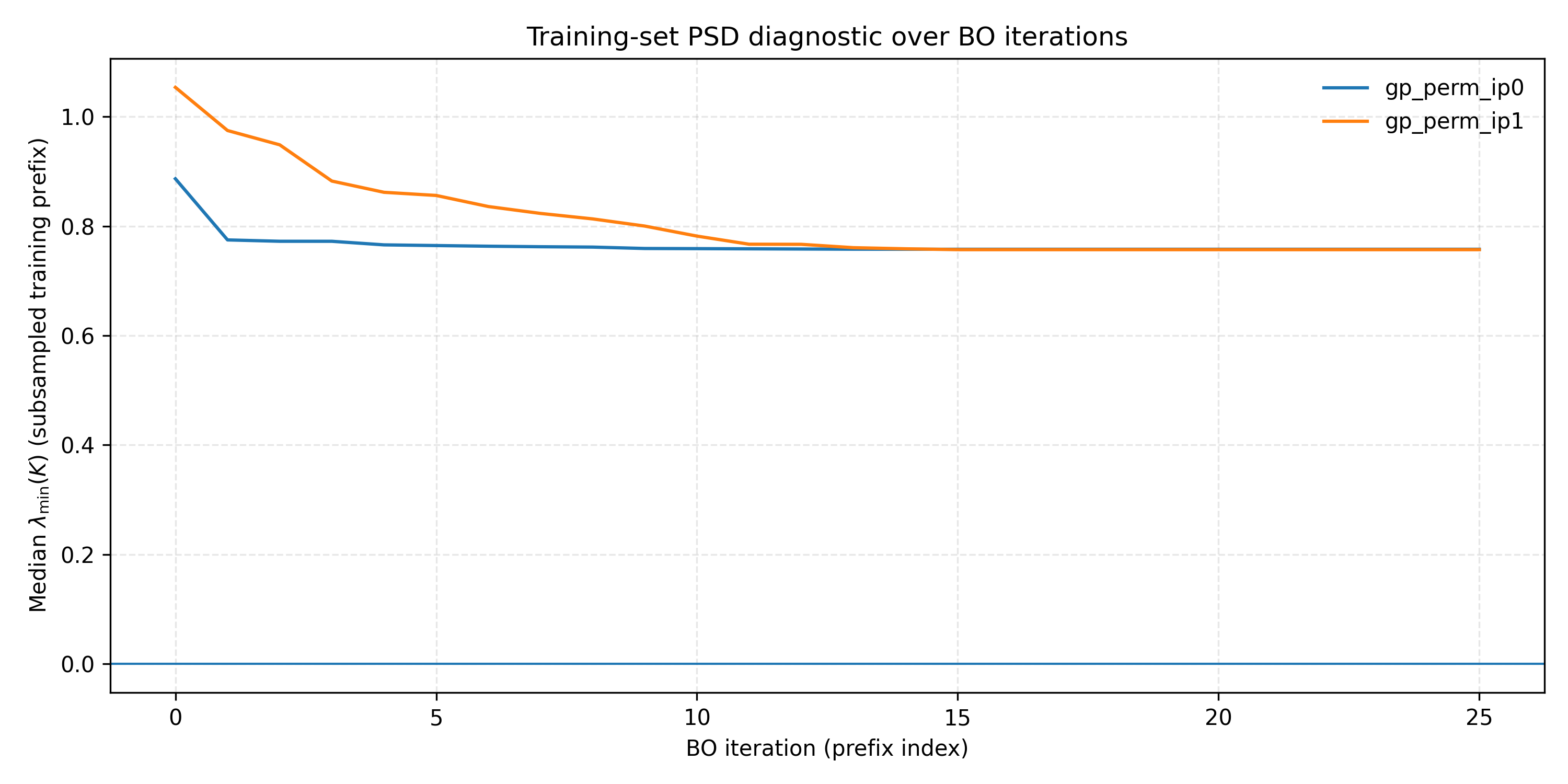}
\caption{Training-set PSD diagnostic during BO: median $\lambda_{\min}(K)$ (subsampled training prefixes) versus BO iteration.
Both GP-Perm variants have positive minimum eigenvalues on every evaluated training prefix.}
\label{fig:appendixD_training_curve}
\end{figure}

We next evaluate PSD behavior on the BO training sets encountered during optimization. For each trial and BO iteration,
we form a subsampled prefix of the training inputs (up to 64 points for efficiency), compute the corresponding kernel matrix,
and record $\lambda_{\min}(K)$. Figure~\ref{fig:appendixD_training_curve} reports the median $\lambda_{\min}(K)$ across trials
as a function of BO iteration.

\begin{table}[t!]
\centering
\small
\caption{Training-set PSD summary across all trials and BO iterations (subsampled prefixes). Reported values are the minimum observed $\lambda_{\min}(K)$, the median over all evaluated prefixes, and the fraction of evaluated prefixes where $\lambda_{\min}(K)<-\delta$.}
\label{tab:appendixD_training_psd}
\begin{tabular}{lcccccc}
\toprule
\textbf{Variant} & $\min\,\lambda_{\min}$ & $\mathrm{median}\,\lambda_{\min}$ & $\mathrm{mean}\,\lambda_{\min}$ &
$\delta=10^{-10}$ & $\delta=10^{-8}$ & $\delta=10^{-6}$\\
\midrule
GP-Perm (-IP) & 0.176 & 0.764 & 0.727 & 0.000 & 0.000 & 0.000 \\
GP-Perm (+IP) & $5.62\times 10^{-7}$ & 0.794 & 0.725 & 0.000 & 0.000 & 0.000 \\
\bottomrule
\end{tabular}
\end{table}

\subsection{Interpretation of the finite-design evidence}
\label{appendixd:interpretation}

Across both offline random designs and BO training prefixes, the raw kernel matrix had $\lambda_{\min}(K)>0$ in every evaluated case. These eigendecompositions are direct deterministic certificates for the particular finite designs and fitted hyperparameters that were evaluated; they are not extrapolated into a universal Mercer claim. The offline medians in Table~\ref{tab:appendixD_offline_psd} indicate substantial typical margins. More strongly, the minima over all evaluated BO prefixes in Table~\ref{tab:appendixD_training_psd} give pathwise raw-kernel margins of $0.176$ for GP-Perm (-IP) and $5.62\times10^{-7}$ for GP-Perm (+IP). Proposition D.4 therefore certifies every symmetric perturbation smaller in spectral norm than the corresponding margin, even before likelihood noise or jitter is added.

Diagonal regularization enlarges these margins exactly:
\begin{equation}
\lambda_{\min}(K+\sigma_n^2I_N+\tau I_N)
=\lambda_{\min}(K)+\sigma_n^2+\tau.
\label{eq:appendixd_shift}
\end{equation}
For example, at the first staged-jitter level $\tau=10^{-5}$, the smallest reported (+IP) training-prefix eigenvalue implies
\begin{equation}
\lambda_{\min}(C)
\geq 5.62\times10^{-7}+10^{-5}
=1.0562\times10^{-5},
\end{equation}
even if the additional positive likelihood variance is omitted from the bound. Larger stages increase this margin further.

The resulting guarantee is deliberately finite and operational. Proposition D.1 states when a universal theorem would follow; Propositions D.2--D.3 show that PSD can be guaranteed on a finite design without that global premise; Proposition D.4 quantifies robustness around the matrices actually encountered. The same arguments cover a single unordered set, two unordered sets, and the full construction with auxiliary and interaction terms. The simulator-free geometry study adds independent evidence that the implemented construction is invariant to float64 precision, varies continuously under coordinate perturbations, distinguishes cross-group alignment through $\mathcal R$, and remains positive definite throughout a declared $\varepsilon$--$\ell$ grid. These kernel facts do not by themselves imply improved BO performance. Together with the retained experiment-specific spectra, however, they explain why GP-Perm can support stable GP inference in the reported runs while avoiding an unsupported claim of unconditional PSD over the unrestricted input space.

\section{Complete sensitivity analysis results}
\label{appendixe}

This appendix reports the benchmark-driven sensitivity analysis used \emph{only} to select GP-Perm and DKL-DS configurations before running CCS case studies. In particular, we introduce here a normalized aggregation score that is used for \emph{variant selection} across heterogeneous synthetic benchmarks. This normalization is \emph{not} used for the main benchmark comparisons between model families (GP-Perm, DKL-DS, and baselines), where results are reported per benchmark using the raw best-so-far trajectories and their benchmark-specific AUC and final best value.

\subsection{Variant selection score (normalized aggregation)}
\label{appendixe:selection_score}

For a given benchmark $e$, trial $r$, and model variant $m$, let $b_t^{(e,r,m)}$ denote the best-so-far value after BO iteration $t$.
We summarize each trial by:
\begin{align}
\mathrm{AUC}^{(e,r,m)} &:= \sum_{t=1}^{T_e} b_t^{(e,r,m)}, \\
b_T^{(e,r,m)} &:= b_{T_e}^{(e,r,m)} ,
\end{align}
where $T_e$ is the BO horizon for benchmark $e$.

Because benchmarks differ in scale, we min--max normalize each metric \emph{within} each benchmark across the compared variants:
\begin{equation}
\widehat{m}_{e}\bigl(m,r\bigr)
=
\frac{
m_{e}\bigl(m,r\bigr)-\min_{m'} m_{e}\bigl(m',\cdot\bigr)
}{
\max_{m'} m_{e}\bigl(m',\cdot\bigr)-\min_{m'} m_{e}\bigl(m',\cdot\bigr)
}
\in [0,1],
\end{equation}
where $m_e$ denotes either $\mathrm{AUC}$ or $b_T$ for benchmark $e$, and the $\min$/$\max$ are taken over all compared variants (pooling trials).

\subsection{Model variants and sensitivity analysis}
\label{appendixe:search_space}

The benchmark stage includes a targeted sensitivity analysis for the two permutation-invariant surrogate families, GP-Perm and DKL-DS. For GP-Perm, we vary only the Sinkhorn-divergence regularization parameter $\varepsilon$, which following \cite{feydy2019interpolating}, controls a continuum between geometric and kernel-based comparisons: as $\varepsilon\!\to\!0$, $S_{\varepsilon}$ approaches the unregularized OT cost (Wasserstein-type matching), while as $\varepsilon\!\to\!\infty$ it converges to an MMD-like discrepancy induced by a Gibbs kernel $k_{\varepsilon}(\bm{s},\bm{t}) \propto \exp\!\big(-C(\bm{s},\bm{t})/\varepsilon\big)$.
For empirical sets with uniform weights, this high-$\varepsilon$ regime yields a double-sum form,
\begin{equation}
\label{eq:mmd_limit}
\mathrm{MMD}^2_{k_\varepsilon}(S,T)
=
\frac{1}{m^2}\sum_{a,a'} k_{\varepsilon}(\bm{s}_a,\bm{s}_{a'})
+\frac{1}{m^2}\sum_{b,b'} k_{\varepsilon}(\bm{t}_b,\bm{t}_{b'})
-\frac{2}{m^2}\sum_{a,b} k_{\varepsilon}(\bm{s}_a,\bm{t}_b),
\end{equation}
thereby linking GP-Perm to classical double-sum set kernels in the large-$\varepsilon$ limit.

For DKL-DS, we evaluate architectural variants that differ primarily in the choice of invariant pooling operators and whether ARD is enabled in the latent GP kernel. The explored variants are summarized in Table~\ref{tab:variant_sweeps}. All remaining hyperparameters are held fixed within each family.

We select one GP-Perm variant and one DKL-DS variant by maximizing the selection score $S(m)$ defined in Section~\ref{appendixe:selection_score}. Detailed sensitivity results are summarized below.

\begin{table}[t!]
\centering
\caption{Model variants explored in the benchmark sensitivity analysis. Left: GP-Perm variants (varying only $\varepsilon$). Right: DKL-DS variants (varying pooling operators).}
\label{tab:variant_sweeps}
\begin{minipage}[t]{0.4\textwidth}
\centering
\subcaption*{\textbf{GP-Perm variants}}
\begin{tabular}{@{}lc@{}}
\toprule
\textbf{Variant} & $\boldsymbol{\varepsilon}$ \\
\midrule
GP-Perm-1  & $10^{-2}$ \\
GP-Perm-2  & $5\cdot 10^{-2}$ \\
GP-Perm-3  & $5\cdot10^{-3}$ \\
GP-Perm-4  & $10^{-4}$ \\
GP-Perm-5  & $0.5$ \\
GP-Perm-6  & $1$ \\
GP-Perm-7  & $5$ \\
GP-Perm-8  & $10$ \\
GP-Perm-9  & $2$ \\
GP-Perm-10 & $0.1$ \\
\bottomrule
\end{tabular}
\end{minipage}
\hfill
\begin{minipage}[t]{0.58\textwidth}
\centering
\subcaption*{\textbf{DKL-DS variants}}
\begin{tabular}{@{}ll@{}}
\toprule
\textbf{Variant} & \textbf{Pooling} \\
\midrule
DKL-DS-1  & mean + max + std + GEM ($p=\{0,2\}$) + ARD \\
DKL-DS-2  & mean + std + ARD \\
DKL-DS-3  & mean + max + LSE + ARD \\
DKL-DS-4  & GEM ($p=\{0,1,2\}$) + ARD \\
DKL-DS-5  & mean \\
DKL-DS-6  & mean + max + std + GEM ($p=\{0,2\}$) \\
DKL-DS-7  & mean + max + std + GEM ($p=\{0,2\}$) + LSE + ARD \\
DKL-DS-8  & mean + max + std + GEM ($p=\{0,2\}$) + LSE \\
\bottomrule
\end{tabular}
\end{minipage}
\end{table}

\subsection{GP-Perm: sensitivity to Sinkhorn regularization}
\label{appendixe:gpperm}

\begin{figure}[t!]
\centering
\includegraphics[width=0.95\linewidth]{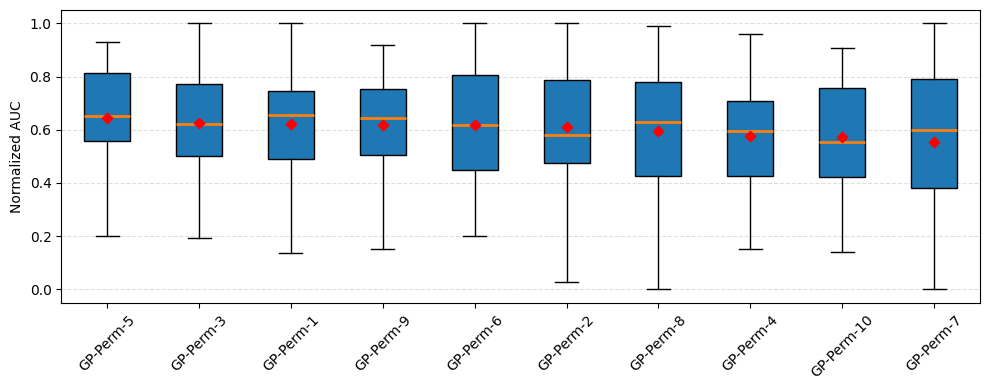}
\caption{GP-Perm sensitivity analysis across benchmarks. Distribution of normalized $\mathrm{AUC}$ over all trials (after per-benchmark normalization).}
\label{fig:appendixE_gpperm_auc}
\end{figure}

\begin{figure}[t!]
\centering
\includegraphics[width=0.95\linewidth]{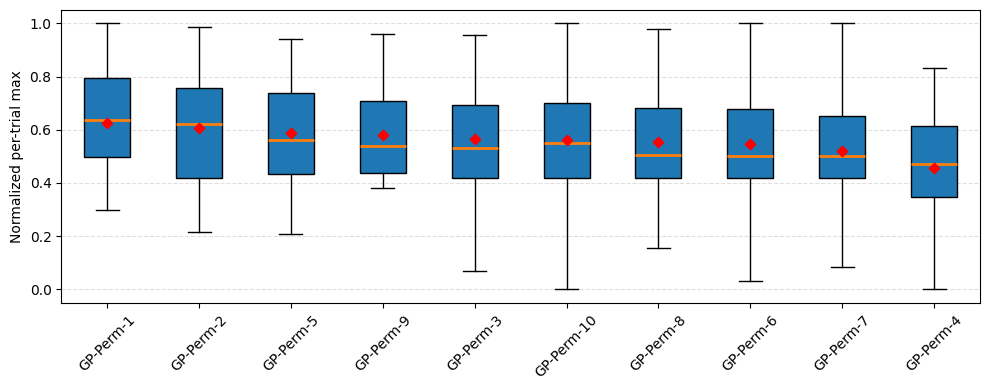}
\caption{GP-Perm sensitivity analysis across benchmarks. Distribution of normalized final best value $b_T$ over all trials (after per-benchmark normalization).}
\label{fig:appendixE_gpperm_max}
\end{figure}

Figures~\ref{fig:appendixE_gpperm_auc}--\ref{fig:appendixE_gpperm_max} summarize GP-Perm sensitivity across benchmarks as distributions over trials of (normalized) $\mathrm{AUC}$ and final best value $b_T$, respectively, after per-benchmark normalization (Section~\ref{appendixe:selection_score}).

\subsection{DKL-DS: sensitivity to pooling design and encoder variants}
\label{appendixe:dklds}

\begin{figure}[t!]
\centering
\includegraphics[width=0.95\linewidth]{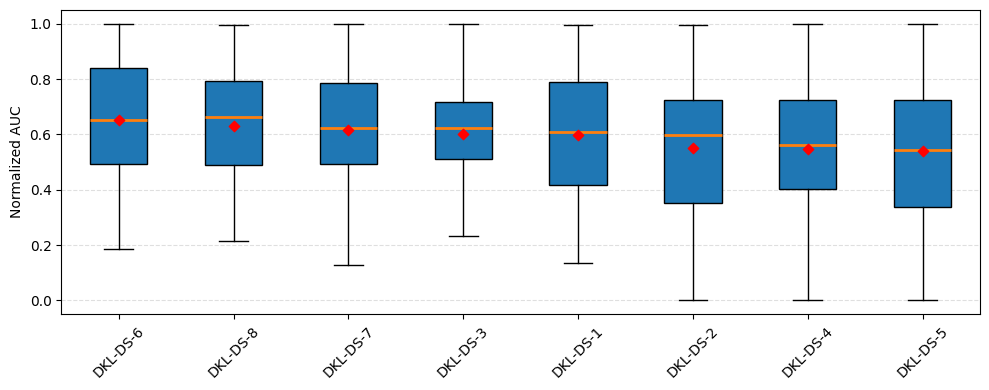}
\caption{DKL-DS sensitivity analysis across benchmarks. Distribution of normalized $\mathrm{AUC}$ over all trials (after per-benchmark normalization).}
\label{fig:appendixE_dklds_auc}
\end{figure}

\begin{figure}[t!]
\centering
\includegraphics[width=0.95\linewidth]{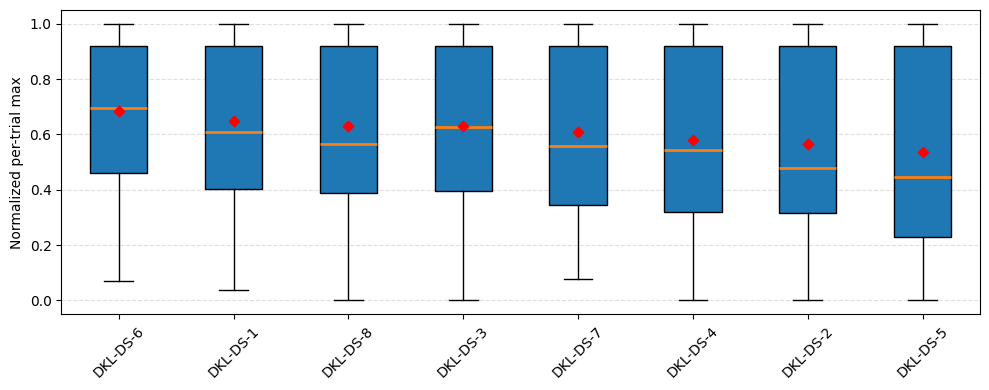}
\caption{DKL-DS sensitivity analysis across benchmarks. Distribution of normalized final best value $b_T$ over all trials (after per-benchmark normalization).}
\label{fig:appendixE_dklds_max}
\end{figure}

Figures~\ref{fig:appendixE_dklds_auc}--\ref{fig:appendixE_dklds_max} summarize DKL-DS sensitivity across benchmarks as distributions over trials of (normalized) $\mathrm{AUC}$ and final best value $b_T$, respectively, after per-benchmark normalization (Section~\ref{appendixe:selection_score}).

\end{document}